\documentclass{article}

\newif\ifappendix
\appendixfalse
\ifappendix
\else
\usepackage{xr}
\makeatletter

\newcommand*{\addFileDependency}[1]{% argument=file name and extension
\typeout{(#1)}% latexmk will find this if $recorder=0
\@addtofilelist{#1}
\IfFileExists{#1}{}{\typeout{No file #1.}}
}\makeatother

\newcommand*{\myexternaldocument}[1]{%
\externaldocument{#1}%
\addFileDependency{#1.tex}%
\addFileDependency{#1.aux}%
}
    \myexternaldocument{appendix}
\fi

\usepackage[final]{corl_2022}
\usepackage{times}

\usepackage[numbers]{natbib}
\usepackage{caption}
\usepackage{subcaption}
\usepackage{wrapfig}
\usepackage{adjustbox}
\usepackage{floatrow}
\newfloatcommand{capbtabbox}{table}[][\FBwidth]
\usepackage{blindtext}
\usepackage{etoc} 
\usepackage{bibunits}

\newcommand{\MA}[0]{{\mathcal{A}}}
\newcommand{\MB}[0]{{\mathcal{B}}}

\newcommand{\PA}[0]{\mathbf{P}_{\MA}}
\newcommand{\PB}[0]{\mathbf{P}_{\MB}}
\newcommand{\NA}[0]{N_{\MA}}
\newcommand{\NB}[0]{N_{\MB}}
\newcommand{\TAB}[0]{\mathbf{T}_{\MA \MB}}

\newcommand{\AtoB}[0]{\MA \to \MB}
\newcommand{\BtoA}[0]{\MB \to \MA}

\newcommand{\VA}[0]{\mathbf{\tilde{V}}_{\MA}}
\newcommand{\VB}[0]{\mathbf{\tilde{V}}_{\MB}}

\usepackage{blindtext}
\usepackage{enumitem,kantlipsum}
\usepackage[font={footnotesize}]{caption}
\usepackage{color, soul} % This allows us to highlight text that you want your co-authors to review
\usepackage[utf8]{inputenc}
\usepackage{graphicx}
\usepackage{multicol}
\usepackage{footmisc}
\usepackage{amssymb}
\usepackage{amsmath}
\usepackage{amsfonts}
\usepackage[ruled]{algorithm}
\usepackage{bbm}
\usepackage{tablefootnote}
\usepackage{multirow}
\usepackage[normalem]{ulem}
\useunder{\uline}{\ul}{}
\usepackage[noend]{algorithmic}
\usepackage{tikz}
\usepackage{enumitem}
\usepackage{float}
\usepackage{hyperref}
\usepackage{array}

\usepackage{booktabs} % for professional tables
\usepackage{mathtools}

\usetikzlibrary{math,shapes,positioning}
\usepackage{listings}

\renewcommand{\vec}[1]{\mathbf{#1}}

\newcommand{\clipart}[1]{\includegraphics[height=6mm]{images/#1.png}}

\newcommand{\cameraready}[1]{{\color{blue} #1}}
\title{TAX-Pose: Task-Specific Cross-Pose Estimation\\ for Robot Manipulation}

\author{
  Chuer Pan\thanks{Equal Contribution. See Appendix \ref{appendix:contributions} for a detailed list of each author's contributions.}, Brian Okorn\footnotemark[1], Harry Zhang\footnotemark[1], Ben Eisner\footnotemark[1], David Held \\
  Robotics Institute, School of Computer Science\\
  Carnegie Mellon University, 
  United States\\
  \texttt{\{chuerp, bokorn, haolunz, baeisner, dheld\}@andrew.cmu.edu}
}

\begin{document}
\maketitle

%===============================================================================

% \begin{bibunit}[corlabbrvnat]
\begin{abstract}
   How do we imbue robots with the ability to efficiently manipulate unseen objects and transfer relevant skills based on demonstrations? End-to-end learning methods often fail to generalize to novel objects or unseen configurations. Instead, we focus on the task-specific pose relationship between relevant parts of interacting objects. We conjecture that this relationship is a generalizable notion of a manipulation task that can transfer to new objects in the same category; examples include the relationship between the pose of a pan relative to an oven or the pose of a mug relative to a mug rack. We call this task-specific pose relationship ``cross-pose" and provide a mathematical definition of this concept. We 
   propose a vision-based system that learns to estimate the cross-pose between two objects for a given manipulation task using learned cross-object correspondences. The estimated cross-pose is then used to guide a downstream motion planner to manipulate the objects into the desired pose relationship (placing a pan into the oven or the mug onto the mug rack). 
   %We train a cross-pose estimator in simulation and we demonstrate the capability of our system to generalize to unseen objects in both simulation and the real world, deploying our policy on a Franka Emika with no finetuning.
   We demonstrate our method's capability to generalize to unseen objects,
   in some cases after training on only 10 demonstrations in the real world.
   %deploying our policy on a Franka Panda 
   Results show that our system achieves state-of-the-art performance in both simulated and real-world experiments across a number of tasks. Supplementary information and videos can be found on our \href{https://sites.google.com/view/tax-pose/home}{project website}.
\end{abstract}


\maketitle
\begin{figure*}[h]
    \centering
    \includegraphics[width=\textwidth, clip]{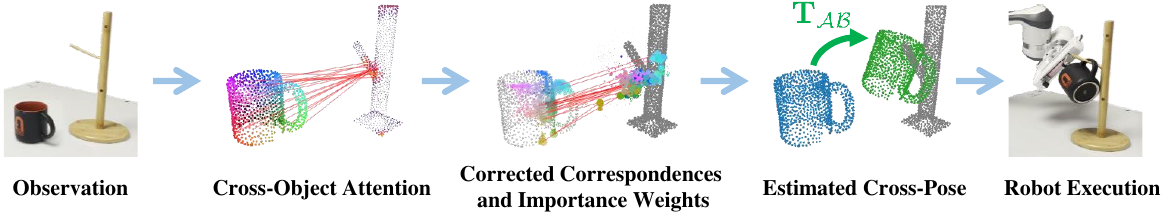}
    \vspace{-20pt}
    \caption{To solve a relative placement task, TAX-Pose uses cross-object attention to estimate dense cross-object correspondences and importance weights for each object point. This dense estimate is mapped to a single ``cross-pose" which the robot uses to accomplish the given task.}
    \label{fig:teaser_pipeline}
\end{figure*}
\section{Introduction}

Many manipulation tasks require a robot to move an object to a location relative to another object.  For example, a cooking robot may need to place a lasagna in an oven, place a pot on a stove, place a plate in a microwave, place a mug onto a mug rack, or place a cup onto a shelf.  Understanding and placing objects in task-specific locations is a key skill for robots operating in human environments. Further, this skill should generalize to novel objects within the training categories, such as placing new trays into the oven or new mugs onto a mug rack.
A common approach in robot learning is to train a policy ``end-to-end," mapping from pixel observations to low-level robot actions.  However, end-to-end trained policies cannot easily reason about complex pose relationships such as the ones described above, and they have difficulty generalizing to unseen object configurations.

In contrast, we propose 
a method that learns to reason about the 3D geometric relationship between a pair of objects.
For the type of tasks defined above, the robot needs to reason about the relationship between key parts on one object with respect to key parts on another object.  For example, to place a mug on a mug rack, the robot must reason about the relationship between the pose of the mug handle and the pose of the mug rack; if the mug rack changes its pose, then the pose of the mug must change accordingly in order to still be placed on the rack (see Figure~\ref{fig:rel_placement}).  We name this task-specific notion of the pose relationship between a pair of objects as ``cross-pose" and we formally define it mathematically.  Further, we propose a vision system that can efficiently estimate the cross-pose from a small number of demonstrations of a given task, generalizing to novel objects within the training categories.
To complete the manipulation task, we use the estimated cross-pose as the target of a motion planning algorithm, which will move the objects into the desired configuration (e.g. placing the mug onto the rack, placing the lasagna into the oven, etc).

\begin{wrapfigure}[25]{r}{0.4\linewidth}
    \centering
    \includegraphics[width=\textwidth, trim={0 0 0 0},clip]{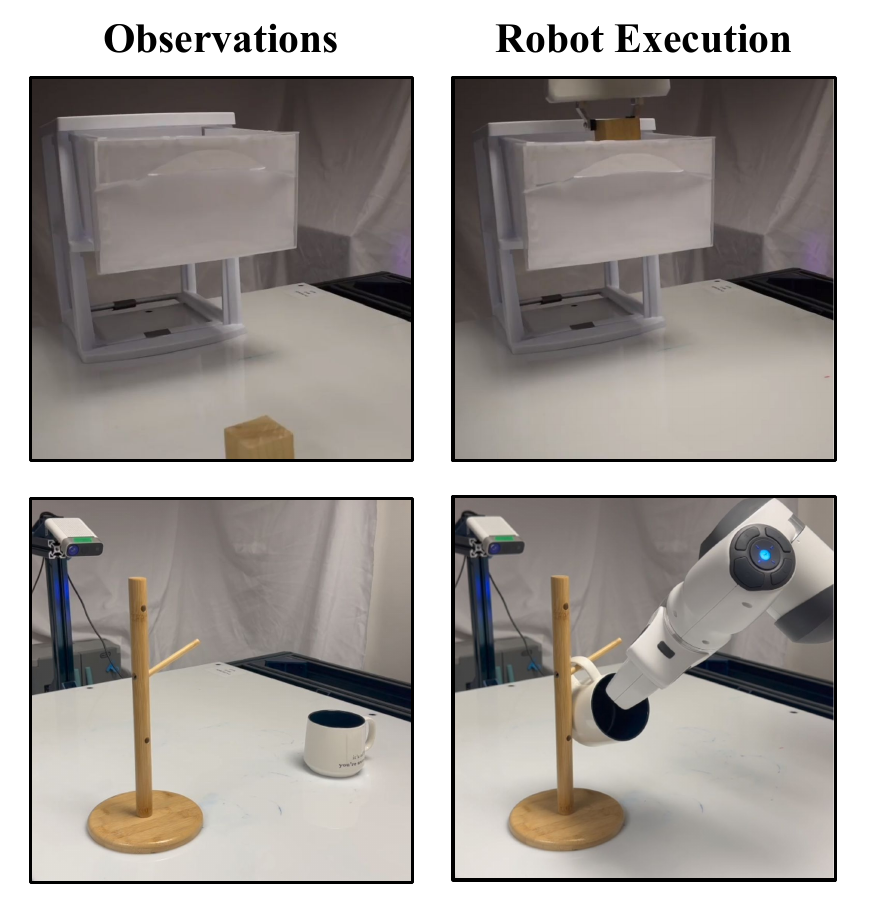}
    \vspace{-10pt}
    \caption{We study relative placement tasks, in which one object needs to be placed in a position relative to another object. Here are two of the tasks that we demonstrate our method on: \textbf{Top:} \textit{PartNet-Mobility Placement Task} requires one object (e.g. a block) to be placed relative to another object (e.g. a drawer) by a semantic goal position (e.g. inside); \textbf{Bottom:} \textit{Mug Hanging Task} requires placing the mug's handle on the mug rack.}
    \label{teaser}
    \vspace{-10pt}
\end{wrapfigure}

In this paper, we present TAX-Pose (TAsk-specific Cross-Pose), a deep 3D vision-based method that learns to predict a task-specific pose relationship between a pair of objects from a set of demonstrations. 
%We use this predicted pose relationship to plan a trajectory that moves the objects into the desired relative pose. 
Our cross-pose estimation system is provably translation equivariant and can generalize from a small number of real-world demonstrations (in some cases as few as 10) to new objects in unseen poses. 

The contributions of this paper include:
\begin{enumerate}[nosep,leftmargin=*]
    % \item A novel goal pose prediction model based on a demonstration from another instance.
    \item A precise definition of ``cross-pose," which defines a task-specific pose relationship between two objects.
    \item A novel method that estimates soft-correspondences between two objects, from which the cross-pose between the objects can be estimated (see Figure~\ref{fig:teaser_pipeline}); this method is provably translation equivariant and can learn from a small number of real-world demonstrations.
    \item  A  robot system to manipulate objects into the desired cross-pose to achieve a given manipulation task. 
\end{enumerate}

We present simulated and real-world experiments to test the performance of our system in achieving a variety of relative placement manipulation tasks. %, learning from a small number of demonstrations. %We show that our model can generalized to novel objects, accurately hanging unseen mugs and interacting with a large variety of placement objects.
We demonstrate our method on 
%We show the generalizabilty of our model though 
a semantic placement task, in which the robot must place an object in, on, or around a novel object (Figure \ref{teaser}, top). We also demonstrate our method on precise placement tasks, such as hanging a mug on a rack (Figure \ref{teaser}, bottom) or placing a bottle or bowl on a shelf; in both cases our method generalizes to new object configurations and new objects within the training categories.

\section{Related Work}

\textbf{Object Pose Estimation}:
Pose estimation is the task of detecting and inferring the 6DoF pose of an object, which includes its position and orientation, with respect to some previously defined object reference frame~\cite{lowe1999object,rothganger20063d,xiang2017posecnn,he2020pvn3d,he2021ffb6d,turpin2021gift}. Recent work~\cite{manuelli2019kpam,qin2020keto, vecerik2021s3k,manuelli2021keypoints, pan2023tax} proposed to use 3D semantic keypoints as an alternative form of object representation. While keypoint-based methods can generalize within an object class, they require a significant amount of hand annotated data or access to a simulated version of the task to learn to estimate the keypoint locations. In contrast, our method is able to learn from just 10 real-world demonstrations. Another approach is to use dense embeddings, such as Dense Object Nets (DON)~\cite{florence2018dense, zhang2016health} and Neural Descriptor Fields (NDF)~\cite{simeonov2021neural, zhang2021robots},
which achieve generalization across classes by predicting dense embeddings in the observation and matching them to embeddings of the demonstration objects. However, DON~\cite{florence2018dense, zhang2023flowbot++} and NDF~\cite{simeonov2021neural, sim2019personalization} assume that the target object is moved relative to a static reference object  in a ``known canonical configuration" (e.g. the pose of the mug rack in NDF~\cite{simeonov2021neural} is assumed to be known and fixed). In contrast, our method reasons about the %``cross-pose" which describes a task-specific
geometric relationship between a pair of objects and hence does not need to assume a static environment. Thus, for example, our method is able to perform the mug hanging task while varying the pose of the mug rack (see our \href{https://sites.google.com/view/tax-pose/home}{project website}), whereas the baselines (DON~\cite{florence2018dense}, NDF~\cite{simeonov2021neural}) cannot. Further, we show that our method significantly outperforms both the DON~\cite{florence2018dense} and NDF~\cite{simeonov2021neural} baselines, especially when given a very small number of demonstrations. 

\textbf{Point Cloud Registration}: Our method for estimating the cross-pose between two objects builds upon previous work in point cloud registration. The typical objective in point cloud registration is to find the optimal rigid alignment between two point clouds, to minimize the sum of squared distances between two sets of points. Traditionally, Iterative Closest Point (ICP)~\cite{besl1992method, zhang2020dex} and its variants~\cite{bouaziz2013sparse, rusinkiewicz2001efficient, segal2009generalized,agamennoni2016point, hinzmann2016collaborative, hahnel2002probabilistic} have been used to compute the optimal rigid alignment between two point clouds.
Deep Closest Point (DCP)~\cite{wang2019deep} avoids local minima common for ICP by seeking to approximate correspondence in a high-dimensional learned feature space.  Our method builds upon the architecture of DCP for cross-pose estimation; however, in contrast to point cloud registration, in which the objective is to minimize the sum of squared distances between two sets of points on the same object in two different poses, our objective is to estimate a task-specific pose relationship between two different objects.
%From a technical standpoint, our work is the first to introduce the notion of cross-pose, which we define as a function of cross-object correspondences in Section~\ref{sec:Problem Statement}. 
Extending the framework from DCP, we learn a residual to the soft correspondences, allowing for points to match outside the convex hull of each object. This component is necessary when computing soft correspondences between objects of drastically different morphologies (such as a mug and a mug rack).  

\section{Problem Statement}
\label{sec:Problem Statement}

\begin{wrapfigure}[16]{R}{0.4\textwidth}
    \begin{center}
    \vspace{-10pt}
    \includegraphics[width=\textwidth, trim={0 0 0 0}, clip]{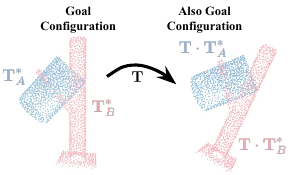}
    \vspace{-20pt}
    \end{center}
    \caption{If we transform both the action object (mug) and the anchor object (rack) by the same transform, then the relative pose between these objects is unchanged (the mug is still ``on" the rack) so the mug is still in the goal configuration. 
    % \dave{It seems that we don't refer to this figure anywhere in the text.  We should figure out where to reference it.  One thought - this figure really shows the definition of a relative placement task, so it should probably be moved to the section where we define relative placement tasks (and referred to there in the text) - although in that case maybe the equations beneath the figures should be removed?}
    }
    \label{fig:rel_placement}
\end{wrapfigure}

\textbf{Relative placement tasks:}
In this paper, we are specifically interested in ``relative placement tasks."   Given two objects, $\mathcal{A}$ and $\mathcal{B}$, a ``relative placement task" is the task of placing object $\mathcal{A}$ at a pose relative to object $\mathcal{B}$. For example, consider the task of placing a lasagna in an oven, placing a mug on a rack, or placing a robot gripper on the rim of a mug.  All of these tasks involve placing one object (which we call the ``action'' object $\mathcal{A}$) at a semantically meaningful location relative to another object (which we call the ``anchor'' object $\mathcal{B}$)\footnote{Note that the definition of action and anchor is symmetric; either object can be treated as the action object and the other as the anchor. }. 
%Intuitively, a relative placement task is a task such that  only the relationship between objects $\MA$ and $\MB$ is important for task success.  

% \ben{I kind of feel we should be more explicit about reference frames here. What we want to say is: `this is the pose of A expressed in the coordinate frame of F'. One way to write this would be $T_{AF}$, which is what Murray-Lee-Sastry do. Therefore you could say, ``Suppose that $\mathbf{T}_{\MA F}^*$ and $\mathbf{T}_{\MB F}^*$ are poses for objects $\MA$ and $\MB$ respectively in some arbitrary reference frame $F$ (the body frame of $B$, for instance, might be a common reference frame to choose as $F$)... Importantly, the body frames of A and B are also arbitrary: we can choose to place the body frame anywhere relative to the mass of the object (centroid is commonly used); what is not arbitrary is the relationship between the objects when described in a single frame (i.e. F).}
%chuer{I kind of echo Ben's views, below is my }
Specifically, suppose that $\mathbf{T}_{\MA}^*$ and $\mathbf{T}_{\MB}^*$ 
are $SE(3)$ poses for objects $\MA$ and $\MB$ respectively (in a shared world reference frame\footnote{All $SE(3)$ transformations in this work are defined in a fixed, arbitrary world frame.}) for which a desired task is considered complete (lasagna is in the oven; mug is on the rack, etc). Then for a relative placement task, if objects $\MA$ and $\MB$ are in poses $\mathbf{T} \cdot \mathbf{T}_{\MA}^*$ and $\mathbf{T} \cdot \mathbf{T}_{\MB}^*$ (respectively) for any transform $\mathbf{T}$, then the task will also be considered to be complete, as seen in Figure~\ref{fig:rel_placement}.  
% In other words, if $\mathbf{T}_{\MB}^*$ represents the pose of the oven and $\mathbf{T}_{\MA}^*$ represents the pose of the lasagna in that oven (at task completion); then if we transform the both the lasagna and oven poses by $\mathbf{T}$, then the lasagna will still be located inside the oven.
In other words, if $\mathbf{T}_{\MB}^*$ represents the pose of the rack and $\mathbf{T}_{\MA}^*$ represents the pose of the mug on the rack (at task completion); then if we transform the both the mug and rack poses by $\mathbf{T}$, then the mug will still be located on the rack.
Formally, this property can be defined with the following Boolean function,
\begin{equation}
    \text{RelPlace}(\mathbf{T_\MA}, \mathbf{T_\MB}) = \textbf{\textsc{Success}}  \text{ iff}~\exists \mathbf{T} \in SE(3) \text{ s.t. } \mathbf{T_{\MA}} = \mathbf{T} \cdot \mathbf{T_{\MA}^{*}} \text{ and } \mathbf{T_{\MB}} = \mathbf{T} \cdot \mathbf{T_{\MB}^{*}}.
\label{eq:relplace_def_simple}
\end{equation}
% \dave{We have formally defined relative placement tasks in Appendix X.} \dave{Or we can add the definition here, it's pretty short}\chuer{addded}
%OLD VERSION:
%We first define the notion of a ``relative placement task." 
%Relative placement tasks can either admit a single unambiguous solution (i.e. placing an asymmetric object in a specific pose, such as inserting an asymmetric peg into a hole), or a set of solutions (e.g. when there are object symmetries, or when the goal is semantically defined, such as ``place the object somewhere on top of the table'').  
For many real semantic placement tasks, there are actually sets of valid solutions which solve each task (i.e., there are many potential locations to place an object on a table to achieve a semantic ``object-on-table'' relationship). However, for this work, we consider precise placement tasks under the simplifying assumption that, for a given pose of object $\mathcal{B}$, there is a single, unambiguous pose of object $\MA$ needed to achieve the task.
 
% In this work, we make the simplifying assumption that, for a given pair of objects $\mathcal{A}$ and $\mathcal{B}$, there is a single, unambiguous relative pose needed to achieve a given task.

%Thus, we treat $\mathcal{A}$ as an ``action'' object, and $\mathcal{B}$ as an ``anchor'' object \footnote{Note that the definition of action and anchor is symmetric.}. 

% \dave{Brian - please edit this}
% \rebuttal{In this work, we consider only placement tasks where there is a single, unambiguous solution for how an object should be placed. In other words, for a given pair objects, we assume there is }

\textbf{Definition of Cross-Pose:}  Given the above definition of a relative placement task, our goal will be to determine how to move object $\MA$ so that it will be in the ``goal pose," which, as described above, is defined relative to the pose of object $\MB$. To achieve this, one option is to estimate the poses of objects $\MA$ and $\MB$ separately and then compute the transformation needed to move object $\MA$ into the goal pose. However, the pose estimate of each object will have errors, and these errors will accumulate when the poses are combined  into the single relative pose needed to reach the goal configuration. 
%Further, training a pose estimator requires pose annotations, whereas we only require a small number of demonstrations.
%combining them to estimate the transformation needed to achieve the goal pose.

Instead of estimating the pose of each object independently, we aim to learn a function  $f(\mathbf{P}_\mathcal{A}, \mathbf{P}_\mathcal{B})$, which takes as input the point clouds $\mathbf{P}_\mathcal{A}$ and $\mathbf{P}_\mathcal{B}$ for both objects $\MA$ and $\MB$ %(respectively)
,
% where $\mathbf{P}_k \in \mathbb{R}^{3\times N_k}$, and $N_k$ denotes the number of 3D points in the point cloud of object $k$. 
 where $\mathbf{P}_{\MA} \in \mathbb{R}^{3\times N_{\MA}}$ and $\mathbf{P}_{\MB} \in \mathbb{R}^{3\times N_{\MB}}$ are 3D point clouds of sizes $N_{\MA}$ and $N_{\MB}$, respectively.
 This function outputs an SE(3) rigid transformation,
% \begin{equation}
$f(\mathbf{P}_\mathcal{A}, \mathbf{P}_\mathcal{B}) = \mathbf{T}_{\mathcal{A}\mathcal{B}}$,
% \end{equation}
where we refer to $\mathbf{T}_{\mathcal{A}\mathcal{B}}$ as the ``cross-pose" between object $\mathcal{A}$ and object $\mathcal{B}$. For notational convenience, we occasional write $f$ as a function of the poses $\mathbf{T}_{\MA}$, $\mathbf{T}_{\MB}$ of point clouds $\PA$ and $\PB$ respectively (with respect to a global reference frame) such that $f(\mathbf{T}_{\MA}$, $\mathbf{T}_{\MB}) := f(\PA, \PB)$. This notational change is to make the transformation math more intuitive; in practice, this function only ever receives point clouds as input.

We will define the cross-pose $\mathbf{T}_{\mathcal{A}\mathcal{B}}$ (below) such that, if we transform  object $\mathcal{A}$ by $\mathbf{T}_{\mathcal{A}\mathcal{B}}$, then object $\MA$ will be in the goal pose relative to object $\MB$ for the relative placement task.  For example,  suppose that $\mathbf{T}_{\MA}^*$ and $\mathbf{T}_{\MB}^*$
are poses for objects $\MA$ and $\MB$, respectively, for which a desired relative placement task is considered complete. 
In this configuration, the cross-pose of these objects would be  $f(\mathbf{T}_{\MA}^*, \mathbf{T}_{\MB}^*) = \mathbf{I}$ where $\mathbf{I}$ is the identity, as object $\MA$ does not need to be moved to complete the task.
% ; in other words, we do not need to move object $\MA$ at all to achieve the task. 
Further, based on the definition of a relative placement task given above, if both objects are transformed by the same transform $\mathbf{T}$, then the objects will still be in the desired relative pose, 
    \begin{equation}
 f(\mathbf{T} \cdot \mathbf{T}_{\MA}^*, \mathbf{T} \cdot \mathbf{T}_{\MB}^*) = f( \mathbf{T}_{\MA}^*, \mathbf{T}_{\MB}^*) = \mathbf{I}
    \label{eq:transform_both}
    \end{equation}
    for any transform $\mathbf{T} \in SE(3)$. Now, let us assume that objects $\MA$ and $\MB$ are not in the goal configuration and have pose $\mathbf{T}_{\MA} = \mathbf{T}_\alpha \cdot \mathbf{T}_{\MA}^*$ and $\mathbf{T}_{\MB} = \mathbf{T}_\beta \cdot \mathbf{T}_{\MB}^*$, respectively, for arbitrary transforms $\mathbf{T}_\alpha$ and $\mathbf{T}_\beta \in SE(3)$. We then define the ``cross-pose" of objects $\MA$ and $\MB$ as:
%and a current ``cross-pose" of
% \begin{equation}
% f(\mathbf{T}_\alpha \mathbf{T}_{\MA}^*,  \mathbf{T}_\beta \mathbf{T}_{\MB}^*):= %\mathbf{T}_\alpha \mathbf{T}_{\MA \MB}^* \mathbf{T}_\beta^{-1} = 
% %\mathbf{T}_\alpha \cdot \mathbf{T}_\beta^{-1}.
% \mathbf{T}_\beta \cdot \mathbf{T}_\alpha^{-1}.
% \label{eq:usability-cross-pose}
% \end{equation}
\begin{equation}
f(\mathbf{T}_{\MA}, \mathbf{T}_{\MB}) = f(\mathbf{T}_\alpha\cdot  \mathbf{T}_{\MA}^*,  \mathbf{T}_\beta \cdot \mathbf{T}_{\MB}^*) = \mathbf{T}_{\MA\MB}:= \mathbf{T}_\beta \cdot \mathbf{T}_\alpha^{-1}.
\label{eq:usability-cross-pose}
\end{equation}
Note that this definition is equivalent to Equation~\ref{eq:transform_both} for the special case of $\mathbf{T}_\alpha = \mathbf{T}_\beta$. 
This definition of cross-pose allows us to move object $\MA$ into the goal configuration, relative to object $\MB$: \begin{equation}
\mathbf{T}_{\mathcal{A}\mathcal{B}} \cdot \mathbf{T}_{\MA} = (\mathbf{T}_\beta \cdot \mathbf{T}_\alpha^{-1}) \cdot (\mathbf{T}_\alpha \cdot \mathbf{T}_{\MA}^*) = \mathbf{T}_\beta \cdot \mathbf{T}_{\MA}^*, 
\end{equation}
satisfying the relative placement condition defined in Equation~\ref{eq:relplace_def_simple} with $\mathbf{T} = \mathbf{T}_\beta$. Alternatively, we could  have instead transformed  object $\MB$ by the inverse of the cross-pose to achieve the task.

\section{Method}
% \input{sec-4-methods}
% This is just a clean scratch page to fiddle with the text of methods.

% \newcommand{\MA}[0]{{\mathcal{A}}}
% \newcommand{\MB}[0]{{\mathcal{B}}}

% \newcommand{\PA}[0]{\mathbf{P}_{\MA}}
% \newcommand{\PB}[0]{\mathbf{P}_{\MB}}
% \newcommand{\NA}[0]{N_{\MA}}
% \newcommand{\NB}[0]{N_{\MB}}
% \newcommand{\TAB}[0]{\mathbf{T}_{\MA \MB}}

% \newcommand{\AtoB}[0]{\MA \to \MB}
% \newcommand{\BtoA}[0]{\MB \to \MA}

% \newcommand{\VA}[0]{\mathbf{\tilde{V}}_{\MA}}
% \newcommand{\VB}[0]{\mathbf{\tilde{V}}_{\MB}}
% \newcommand{\VA}[0]{\mathbf{\tilde{V}}_{\mathcal{A}\to \mathcal{B}}}
% \newcommand{\VB}[0]{\mathbf{\tilde{V}}_{\mathcal{B}\to \mathcal{A}}}

\textbf{Overview:}
We frame the task of cross-pose estimation as a soft correspondence-prediction task between a pair of point clouds, followed by an analytical least-squares optimization to find the optimal \textit{cross-pose} for the predicted correspondences. As described in Appendix \ref{appendix:proof}, this correspondence-based approach allows our method to be translation-equivariant: translating either object ($\mathcal{A}$ or $\mathcal{B}$) will lead to a translated cross-pose prediction.
%We will show that this framing allows our cross-pose estimator to be provably translation-equivariant, 
This allows our method to automatically adapt to novel positions of both the anchor and action objects, unlike previous work which assumes a static anchor~\cite{simeonov2021neural}. %Under this framing, we present TAX-Pose, a model that estimates the task-specific cross-pose from a learned soft correspondence prediction. 
Our method for task-specific cross-pose estimation, known as TAX-Pose, consists of the following steps, as shown in Figure~\ref{fig:gc-ipm}:

\begin{figure*}[b]
    \centering
    \includegraphics[width=\textwidth, clip]{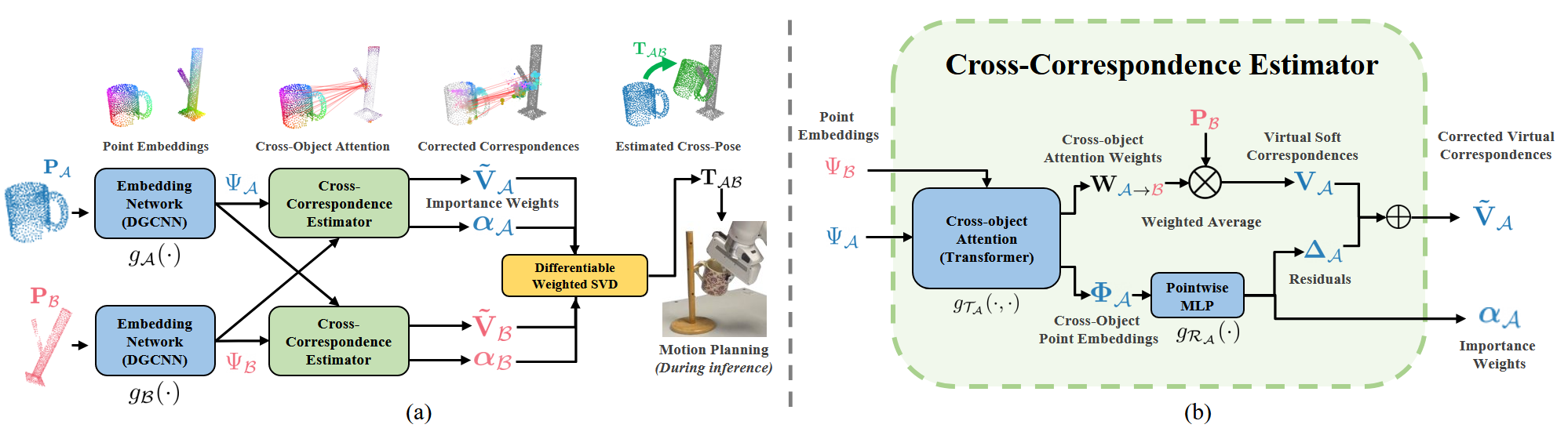}
    \vspace{-20pt}
    \caption{TAX-Pose Training Overview: Given a specific task, our method takes as  input two point clouds and outputs the cross-pose between them needed to achieve the task. TAX-Pose first learns point clouds features using two DGCNN~\cite{phan2018dgcnn} networks and two Transformers~\cite{vaswani2017attention}. Then the learned features are each input to a point residual network to predict per-point soft correspondences and weights across the two objects. The desired cross-pose can be inferred analytically from these correspondences using singular value decomposition.}
    \label{fig:gc-ipm}
\end{figure*}

\begin{enumerate}[leftmargin=*]
    \item \textbf{Soft Correspondence Prediction}: For a pair of objects $\mathcal{A, B}$, a neural network learns to predict a per-point embedding to establish a (soft) correspondence between $\mathcal{A}$ and $\mathcal{B}$, which are called ``virtual soft correspondences." The corresponding points are constrained to be within the convex hulls of $\mathcal{B}$ and $\mathcal{A}$ respectively.
    \item \textbf{Adjustment via Correspondence Residuals}: 
For most estimation tasks,  some points in object $\mathcal{A}$ may not be within the convex hull of object $\mathcal{B}$;
%in direct contact with any point in object $\mathcal{B}$; 
for instance, when a mug is placed on a mug rack, most points on the mug will be outside of the convex hull of the mug rack.
%not be in direct contact with any point on the rack.  
To accommodate these cases, we apply a pointwise residual vector to displace each of the predicted soft correspondences. These ``corrected virtual correspondences" allow points in $\mathcal{A}$ to correspond to locations in free space near $\mathcal{B}$.

    \item \textbf{Find the Optimal Transform}: Because the cross-pose is defined as a rigid transformation of object $\MA$, we use a differentiable weighted SVD to find the transformation that minimizes the weighted least squares difference to the corrected virtual correspondences.
\end{enumerate}

Because each step above is differentiable, the whole model can be optimized end-to-end, despite having an interpretable internal structure which we describe below.
%our method can learn arbitrary correspondences that solve arbitrary cross-pose estimation tasks. 
Our method is heavily inspired by Deep Closest Point (DCP) \cite{wang2019deep}. The key difference between our pose alignment model and DCP is that we predict the cross-pose between two \emph{different} objects for a given task instead of registering two point clouds of an identical object. Additionally, TAX-Pose can predict relationships where these clouds may not have any points of contact or overlap. 

We now describe our cross-pose estimation algorithm in detail. To recap the problem statement, given objects $\MA$ and $\MB$ with point cloud observations $\PA$ $\in \mathbb{R}^{3\times N_{\mathcal{A}}}$, $\PB$ $\in \mathbb{R}^{3\times N_{\mathcal{B}}}$  respectively, our objective is to estimate 
the task-specific cross-pose $\TAB = f(\PA, \PB) \in SE(3)$. Note that the cross-pose between object $\MA$ and $\MB$ is defined with respect to a given task (e.g. putting a lasagna in the oven, putting a mug on the rack, etc).

\subsection{Cross-Pose Estimation via Soft Correspondence Prediction}
\label{sec:soft_corr}
\textbf{Soft Correspondence Prediction}: The first step of the method is to compute two sets of correspondences between $\MA$ and $\MB$, one which maps from points in $\MA$ to $\MB$, and one which maps from points in $\MB$ to $\MA$. These need not be a bijection, and can be asymmetric. 
As we want each step to be differentiable, we 
follow DCP's conventions and estimate a \textit{soft correspondence}. This assigns a \textit{virtual soft corresponding point} $\mathbf{v}_{i}^{\mathcal{A}}\in \mathbf{V}_{\mathcal{A}}$ to every point $\mathbf{p}_{i}^{\MA} \in \PA$ by computing a convex combination of points in $\PB$, and vice versa. Formally:\\
\begin{subequations}
\begin{minipage}{.5\textwidth}
    \begin{equation} 
        \mathbf{v}_{i}^{\mathcal{A}} = \PB \mathbf{w}_{i}^{\AtoB}  \quad \text{s.t.} \quad \sum_{j=1}^{N_{\MB}} w_{ij}^{\AtoB} = 1
    \end{equation}
\end{minipage}%
\hfill
\begin{minipage}{.5\textwidth}
    \begin{equation} 
        \mathbf{v}_{i}^{\mathcal{B}} = \PA \mathbf{w}_{i}^{\BtoA}  \quad \text{s.t.} \quad \sum_{j=1}^{N_{\MA}} w_{ij}^{\BtoA} = 1    
        % \label{soft_correspondence}
        \end{equation}
\end{minipage}
\label{soft_correspondence}
\end{subequations}
with normalized weight vectors $\mathbf{w}_{i}^{\AtoB} \in \mathbf{W}_{\AtoB}$ and $\mathbf{w}_{i}^{\BtoA} \in \mathbf{W}_{\BtoA}$. Importantly, these virtual corresponding points are not constrained to the surfaces of $\MA$ or $\MB$; instead, they are constrained to the convex hulls of $\PB$ and $\PA$, respectively. 

To compute the weights %$\mathbf{W}_{\AtoB}$, $\mathbf{W}_{\BtoA}$ 
$\mathbf{w}_{i}^{\AtoB}$, $\mathbf{w}_{i}^{\BtoA}$  in Equations~\ref{soft_correspondence}a and~\ref{soft_correspondence}b, we first encode each point cloud $\PA$ and $\PB$ into a latent space using a neural network encoder, DGCNN~\cite{phan2018dgcnn}. This encoder head is comprised of two distinct encoders $g_{\MA}$ and $g_{\MB}$, each of which receives point cloud $\PA$ and $\PB$, respectively, zero-centers them, and outputs a dense, point-wise embedding for each object (see Figure~\ref{fig:gc-ipm}):
%\begin{equation}
     $\boldsymbol{\Psi}_\mathcal{A} = g_{\mathcal{A}}(\bar{\mathbf{P}}_{\mathcal{A}}) \in \mathbb{R}^{N_{\mathcal{A}}\times d},\;\;
     \boldsymbol{\Psi}_\mathcal{B} = g_{\mathcal{B}}(\bar{\mathbf{P}}_{\mathcal{B}})\in \mathbb{R}^{N_{\mathcal{B}}\times d}$
     %$\Phi_\mathcal{A} = g_{\mathcal{A}}(\mathbf{P}_\mathcal{A}),\;\;
     %\Phi_\mathcal{B} = g_{\mathcal{B}}(\mathbf{P}_\mathcal{B})$
% \end{equation}
where $\boldsymbol{\psi}_i^{\mathcal{K}} \in \boldsymbol{\Psi}_\mathcal{K}$ is the $d$-dimensional embedding of the $i$-th point in object $\mathcal{K}$, and $\bar{\mathbf{P}}_{\mathcal{K}}$
%:=\mathbf{P}_{\mathcal{K}} - \frac{1}{N_\mathcal{K}} \sum_{i=1}^{N_k} \mathbf{P}_\mathcal{K} \in \mathbb{R}^{3 \times N_{\mathcal{K}}}$ 
is the zero-centered point cloud for object $\mathcal{K}$. 
Because we want the cross-correspondence to incorporate information about both point clouds, we then employ a cross-object attention module between the two dense feature sets 
to obtain \textit{cross-object point embeddings},
$\boldsymbol{\Phi}_{\mathcal{A}} \in \mathbb{R}^{N_\mathcal{A}\times d}$ and $\boldsymbol{\Phi}_{\mathcal{B}} \in \mathbb{R}^{N_\mathcal{B}\times d}$, defined as: 
\begin{equation}
    \boldsymbol{\Phi}_{\mathcal{A}} = \boldsymbol{\Psi}_{\mathcal{A}} + g_{\mathcal{T}_{\mathcal{A}}}(\boldsymbol{\Psi}_{\mathcal{A}}, \boldsymbol{\Psi}_{\mathcal{B}}) , \;\; 
    \boldsymbol{\Phi}_{\mathcal{B}} = \boldsymbol{\Psi}_{\mathcal{B}} + g_{\mathcal{T}_{\mathcal{B}}}(\boldsymbol{\Psi}_{\mathcal{B}}, \boldsymbol{\Psi}_{\mathcal{A}})
\end{equation}
where $g_{\mathcal{T}_{\mathcal{A}}}$,
$g_{\mathcal{T}_{\mathcal{B}}}$ are Transformers~\cite{vaswani2017attention}.

Finally, %to map our estimated features to a set of normalized weights 
recall that our goal was to compute a set of normalized weight vectors $\mathbf{W}_{\AtoB}$, $\mathbf{W}_{\BtoA}$. To compute the virtual corresponding point $\mathbf{v}_i^{\MA}$ assigned to any point $\mathbf{p}_i^{\MA}\in \mathbf{P}^{\MA}$, we can extract the desired normalized weight vector $\mathbf{w}_{i}^{\AtoB}$ from intermediate attention features of the cross-object attention module as:
\begin{equation}
    \mathbf{w}_{i}^{\AtoB} = \text{softmax}\left(\frac{\mathbf{K_{\MB}}\mathbf{q}_{i}^{\MA}}{\sqrt{d}}\right), \;\;
     \mathbf{w}_{i}^{\BtoA} = \text{softmax}\left(\frac{\mathbf{K_{\MA}}\mathbf{q}_{i}^{\MB}}{\sqrt{d}}\right) \;\;
\end{equation}
where $\mathbf{q}_i^{\mathcal{K}} \in \mathbf{Q_{\mathcal{K}}}$, and $\mathbf{Q_{\mathcal{K}}}, \mathbf{K_{\mathcal{K}}} \in \mathbb{R}^{\mathbf{N}_{\mathcal{K}}\times d}$ are the query and key values (respectively) for object $\mathcal{K}$ associated with cross-object attention Transformer module $g_{\mathcal{T}_{\mathcal{K}}}$ (see Appendix \ref{appendix:cross-object-weight}~for details). These weights are then used to compute the virtual soft correspondences $\mathbf{V}_{\mathcal{A}}$, $\mathbf{V}_{\mathcal{B}}$ using Equation\cameraready{~\ref{soft_correspondence}.}

\textbf{Adjustment via Correspondence Residuals:} 
As previously stated, the virtual soft correspondences  $\mathbf{V}_{\mathcal{A}}, \mathbf{V}_{\mathcal{B}}$ given by Equations~\ref{soft_correspondence}a and~\ref{soft_correspondence}b are constrained to be within the convex hull of each object. However, many relative placement tasks cannot be solved perfectly with this constraint.
% correspondences which are constrained to the convex hull are insufficient to express 
For instance, we might want a point on the handle of a teapot to correspond to some point above a stovetop (which lies outside the convex hull of the points on the stovetop). To allow for such off-object correspondences, we further learn a \textit{residual vector}, $\boldsymbol{\delta}_{i}^{\mathcal{A}} \in \boldsymbol{\Delta}_{\mathcal{A}}$ for each 
%point-wise embedding $\boldsymbol{\phi}_i$ 
point $i$
that corrects each virtual corresponding point $\mathbf{v}_{i}^{\mathcal{A}}$. This allows us to displace each virtual corresponding point to any arbitrary location that might be suitable for the task. To compute these residual vectors, we use a point-wise neural network $g_{\mathcal{R}_\MA}$, $g_{\mathcal{R}_\MB}$ to map each point's embedding into a 3D residual vector:
\begin{equation*}
    \boldsymbol{\delta}_i^{\mathcal{A}} = g_{\mathcal{R}_\MA} \left(\boldsymbol{\phi}_i^{\MA}\right) \in \mathbb{R}^3,\;\; 
     \boldsymbol{\delta}_i^{\mathcal{B}} = g_{\mathcal{R}_\MB} \left(\boldsymbol{\phi}_i^{\MB}\right) \in \mathbb{R}^3
\end{equation*}
Applying these residual offsets to the virtual points, we get a set of \textit{corrected virtual correspondences}, $\tilde{\mathbf{v}}_i^{\mathcal{A}} \in \VA$ and $\tilde{\mathbf{v}}_i^{\mathcal{B}} \in \VB$,  defined as
%\cameraready{$\VA = \begin{bmatrix} \tilde{v}_1^{\AtoB} \dots \tilde{v}_{\NA}^{\AtoB} \end{bmatrix}^\top$ and $\VB =  \begin{bmatrix} \tilde{v}_1^{\BtoA} \dots \tilde{v}_{\NB}^{\BtoA} \end{bmatrix}^\top$, where $\tilde{v}_i^{\AtoB}$ and $\tilde{v}_i^{\BtoA}$ are defined as,}
\begin{equation}
    \tilde{\mathbf{v}}_i^{\mathcal{A}} = \mathbf{v}_i^{\mathcal{A}} + \boldsymbol{\delta}_i^{\mathcal{A}},\;\; 
    \tilde{\mathbf{v}}_i^{\mathcal{B}} = \mathbf{v}_i^{\mathcal{B}} + \boldsymbol{\delta}_i^{\mathcal{B}} 
\end{equation}
These corrected virtual correspondences $\tilde{\mathbf{v}}_i^{\mathcal{A}}$ define the estimated goal location relative to object $\MB$ for each point $\mathbf{p}_i \in \PA$ of object $\MA$, and likewise for each point in object $\MB$ (see visualization in Appendix~\ref{appendix:illustration}).

\textbf{Least-Squares Cross-Pose Optimization with Weighted SVD:}
Given the sets of dense correspondences, $\left(\PA, \VA\right)$ and $\left(\PB, \VB\right)$, we would like to compute a single rigid transformation for object $\MA$.  To do so, we solve for the transformation $\TAB$ (the cross-pose) that minimizes the weighted distance between each point and its corrected virtual correspondence.
% For point cloud $\PA$, we have now computed the set of corrected virtual correspondences $\VA$, which is the network's estimate of where each point in $\PA$ would need to move to in order to be in the goal position relative to object $B$ (see Figure S1 in Appendix E.1.).  Likewise, for point cloud $\PB$, we have computed the set of corrected virtual correspondences $\VB$. However, because $\PA$ is assumed to be a rigid object, it cannot independently move each point to the corresponding location in $\VA$.  Instead, we solve for a rigid transformation $\TAB$ (the cross-pose) that minimizes the weighted distance between points and their correspondences. 
Formally, this leads to the following weighted least squares optimization:
\begin{equation}
    \mathcal{J}(\TAB) = \sum_{i=1}^{\NA} \alpha_i^{\MA} || \TAB~\mathbf{p}_{i}^{\MA} - \tilde{\mathbf{v}}_i^{\MA} ||_2^2 + \sum_{i=1}^{\NB} \alpha_i^{\MB} || \TAB^{-1}~\mathbf{p}_{i}^{\MB} - \tilde{\mathbf{v}}_i^{\MB} ||_2^2
    \label{eq:weighted_least_squares}
\end{equation}
where the weights $\alpha_i^{\MA} \in \boldsymbol{\alpha}_{\MA}$, $\alpha_i^{\MB} \in \boldsymbol{\alpha}_{\MB}$ signify the importance of each correspondence and are predicted by a point-wise MLP as shown in Figure~\ref{fig:gc-ipm}. These weights are 
learned end-to-end as parameters of our network; they are visualized in Appendix \ref{appendix:importance_weights},
%the third stage of Figure~\ref{fig:gc-ipm},
which shows that the network has learned to assign more weight to the parts of the object that are most important for the task, such as the region around the mug handle (on the mug) and the region around the peg (on the rack). Equation~\ref{eq:weighted_least_squares} is the well-known weighted Procrustes problem, for which there exists an analytical solution. %As shown in Figure~\ref{fig:gc-ipm}, 
To maintain the differentiablity of the system, we use a weighted differentiable SVD operation~\cite{papadopoulo2000estimating} to compute the cross-pose $\TAB$ that minimizes this objective
(see Appendix \ref{appendix:weighted-svd}~for details). This allows us to train the system end-to-end as described below. 

\subsection{TAX-Pose Training Pipeline}

To train our model, we use a segmented set of demonstration point clouds of a pair of objects in the goal configuration. For each demonstration point cloud, we generate multiple training examples by  transforming each object's point cloud, $\PA$ and $\PB$ by random SE(3) transformations $\mathbf{T}_\alpha$ and $\mathbf{T}_\beta$, respectively. The predicted cross-pose, $\TAB$, is then compared with the ground truth cross-pose, $\TAB^{GT} :=\mathbf{T}_{\beta}\mathbf{T}_{\alpha}^{-1}$, using an average distance loss~\cite{hinterstoisser2012model} with dense regularization (see more details on our training losses in Appendix \ref{appendix:training-supervision}).

\section{Experiments}
To evaluate TAX-Pose, we conduct a wide range of simulated and real-world experiments on two classes of relative placement tasks: NDF~\cite{simeonov2021neural} Tasks and PartNet-Mobility  Placement Tasks. All tasks involve placing an ``action" object at a specific location relative to an anchor object, in which the relative pose is specified by a set of demonstrations. Our method then generalizes to perform this task on novel objects in unseen configurations.
%The Object Placement objective is to place an action object on a flat surface on or near an anchor object. The Mug Hanging task objective is to grasp and then hang unseen mugs on a rack. 
We refer the reader to our \href{https://sites.google.com/view/tax-pose/home}{project website} for additional results and videos.

\subsection{NDF Tasks}
\label{sec:NDF Tasks}
\looseness=-1
We evaluate our method on all three NDF~\cite{simeonov2021neural, devgon2020orienting} tasks (\textit{mug} hanging, \textit{bottle} placement, and \textit{bowl} placement); see Appendix \ref{appendix:results-ndf-bowlbottle} for results on \textit{bottle} and \textit{bowl} placement. Results on \textit{mug} hanging are described in more detail below.

\textbf{Simulation Experiments:}
For our simulation experiments, we perform the task of hanging a mug on a rack as two sequential cross-pose estimation  steps: grasping the mug (estimating the cross-pose between the gripper and the mug) and hanging the mug on the rack (estimating the cross-pose between the mug and the rack). 
In Pybullet \cite{coumans2020, avigal20206, avigal2021avplug}, we simulate a Franka Panda above a table with 4 depth cameras placed on the corners of the table. The model is trained on 10 simulated demonstrations of mug hanging. We evaluate task execution success on unseen mug instances in randomly generated initial configurations.
% To successfully execute the manipulation task of hanging a mug on a rack by the mug's handle requires the successful inference of two sequential task-specific cross-poses: 1) predict a successful grasp pose; 2) predict the hanging pose of mug relative to the rack. 
% In the first stage, it requires our model to reason about the cross-pose between the gripper and the mug, while the second stage requires prediction of the cross-pose between the mug and the rack. We utilize Pybullet \cite{coumans2016pybullet} and simulate a Franka Panda arm situated above a table with 4 depth cameras placed at each table corner. For training, the model is provided with 10 demonstrations in simulation, each on a different mug instances. At test time, we measure the task execution success on unseen mug instances, with randomly generated initial poses. 
% To evaluate robustness to different initial poses, we evaluate on two sets of initial poses: 1) \textit{Upright Pose}: where the mug is initialized to have an upright orientation, and placed randomly on the surface of the table; 2) \textit{Arbitrary Pose}: where the mug is initialized to have arbitrary orientation and position, irrespective of table surface is.
We measure task success rates of 1) \textit{Grasping}, where success is achieved when the object is grasped stably; 2) \textit{Placing}, where success is achieved when the mug is placed stably on the rack; 3)  \textit{Overall}, when the predicted transforms enable both grasp and place success in sequence. We compare our method to Neural Descriptor Field (NDF)~\cite{simeonov2021neural} and Dense Object Nets (DON)~\cite{florence2018dense}. Details of these methods can be found in prior work~\cite{simeonov2021neural, zhang2021robots}. 
%Dense Object Nets (DON)~\cite{florence2018dense} and Neural Descriptor Fields (NDF)~\cite{simeonov2021neural}.

% \underline{Dense Object Nets (DON)}~\cite{florence2018dense}: Using manually labeled semantic keypoints on the demonstration clouds, DON is used to compute sparse correspondences with the test objects. These correspondences are converted to a pose using SVD. A full description can be found in~\cite{simeonov2021neural}.

% \underline{Neural Descriptor Field (NDF)}~\cite{simeonov2021neural}: Using the learned descriptor field for the mug, the positions of a constellation of task specific quarry points are optimized to best match the demonstration using gradient descent.
\textbf{Simulation Results:}
We evaluate our method in simulation in 100 trials consisting of unseen \textit{mug} instances in random initial and goal configurations for both \textbf{Upright} and \textbf{Arbitrary} poses. 
%We compare the performance of our method against DON~\cite{florence2018dense} and NDF~\cite{simeonov2021neural}. 
As shown in Table~1, our method significantly outperforms the baselines 
%See Table~2 for results 
for simulated mug hanging. We report  additional results for simulated \textit{bottle} and \textit{bowl} placement tasks in Table \ref{tab:bottle_bowl} in Appendix \ref{appendix:results-ndf-bowlbottle}.

\looseness=-1
\textbf{Ablation Analysis: } \textit{Effects of Number of Demonstrations.} To study how the number of demonstrations observed affects our method's performance, we train our model on \{10, 5, 1\} demonstrations of upright pose mug hanging. Results are found in Table~2. Our method outperforms the baselines for all number of demonstrations; TAX-Pose can perform well even with as few as 5 demonstrations. 

\textit{Cross-Pose Estimation Design Choices.} We analyze the effects of design choices made in our Cross-Pose estimation algorithm for the upright pose mug hanging task. Specifically, we analyze the effects of 1) computing residual correspondence; 2) the use of \textit{weighted} SVD over non-weighted in computing cross-pose; 3) using a transformer as the cross-object attention, as opposed to simpler model such as a 3-layer MLP.  Table~3 shows that each major component of our system is important for task success. See more ablation experiments in Appendix \ref{appendix:results-ndf-ablations}.

% \begin{table}[ht]
% \begin{varwidth}[b]{0.1\linewidth}
%     \centering
%       \renewcommand{\arraystretch}{1}
%     %  \resizebox{6\linewidth}{!}{
%     \begin{tabular}{c|c c c}
%     % \hline
%     \textbf{Model} & \multicolumn{3}{c} \#\textbf{Demos Used }\\
%     \hline
%           & 1 & 5 & 10\\
%          \hline
%          DON \cite{florence2018dense}  & 0.32 &0.36 & 0.45\\
%          NDF \cite{simeonov2021neural} &  0.46 & 0.70 & 0.88 \\
%           TAX-Pose (Ours) & \textbf{0.77}& \textbf{0.90}& \textbf{0.96}\\
         
%     \hline
%     \end{tabular}
%  % }
%     \caption{Analysis of Number of Demonstrations Used in Training on Upright Pose Mug Hanging Task Success}
%      \label{tab:mug_on_rack_few_shot}
% \end{varwidth}%
% \qquad
% \begin{varwidth}[b]{}
%     \centering
%       
%      
%     \begin{tabular}{c|c c c}
%     \hline
%     \textbf{Ablation} & \textbf{Grasp} & \textbf{Place} & \textbf{Overall}\\
%          \hline
%          No Residual Correspondence  & 0.97 &0.96 & 0.93\\
%          Unweighted SVD &  0.92 & 0.94 & 0.88 \\
%          3-Layer MLP In Place of Attention &  0.90 & 0.82 & 0.76 \\
%          TAX-Pose (Ours)  & \textbf{0.99}& \textbf{0.97}& \textbf{0.96}\\
%     \hline
%     \end{tabular}
%      
%     \caption{Mug Hanging Ablations Results}
%     \label{tab:mug_rack_ablation}
% \end{varwidth}%

% \end{table}

% \begin{table}[htb]

    \noindent \begin{minipage}[t]{.45\textwidth}
    % \vspace{0.5em}
    
    \label{tab:mug_on_rack_exp}
    % \small
        \centering
        \resizebox{\textwidth}{!}{
%         \begin{tabular}{c| c c c}
%     \hline
%   \textbf{Ablation Experiment} & \textbf{Grasp} & \textbf{Place} & \textbf{Overall}\\
  
%     %  & \multicolumn{3}{c}{\textit{Avg Pairwise Angular Error $< 1^{\circ}$}}\\
   
%         % \textbf{Upright Pose} & & &\\
%          \hline
%         % \textit{Loss}&&&\\
%           No $\mathcal{L}_\mathrm{disp}$ &  0.01 & 0 & 0 \\
%           No $ \mathcal{L}_\mathrm{corr}$ &  0.89 & 0.91 & 0.84 \\
%           No $  \mathcal{L}_\mathrm{cons}$ &  \textbf{0.99} & 0.95 & 0.94 \\
%             Scaled Combination: 
%             %of  $ \mathcal{L}_\mathrm{corr} \&   \mathcal{L}_\mathrm{cons}$ (
%             $ 1.1\mathcal{L}_\mathrm{cons} + \mathcal{L}_\mathrm{corr}$ &  0.10 & 0.01 & 0.01 \\
%          \hline 
%           No Adjustment via Correspondence Residuals  & 0.97 &0.96 & 0.93\\
%          \hline 
%          Unweighted SVD &  0.92 & 0.94 & 0.88 \\
%         \hline 
%          No Finetuning for Embedding Network &  0.98 & 0.93 & 0.91 \\
%          No Pretraining for Embedding Network &  \textbf{0.99} & 0.72 & 0.71 \\
%          \hline 
%          3-Layer MLP In Place of Transformer &  0.90 & 0.82 & 0.76 \\
%         \hline
%         Embedding Network Feature Dim = 16 & 0.98 & 0.59 & 0.57 \\
    
%         \hline
%         \hline
%         \textbf{TAX-Pose (Ours)}  & \textbf{0.99}& \textbf{0.97}& \textbf{0.96}\\
%     \hline
%     \end{tabular}

%\begin{table}[h]
\begin{tabular}{c|c c c| c c c}
    \hline
     & \textbf{Grasp} & \textbf{Place} & \textbf{Overall} & \textbf{Grasp} & \textbf{Place} & \textbf{Overall}\\
    % & Grasp & Place & Overall\\
    \hline
    %  & \multicolumn{3}{c}{\textit{Avg Pairwise Angular Error $< 1^{\circ}$}}\\
     &\multicolumn{3}{c|}{\textbf{Upright Pose}}& \multicolumn{3}{c}{\textbf{Arbitrary Pose}} \\
         \hline
         DON \cite{florence2018dense} & 0.91 &0.50 & 0.45 & 0.35 & 0.45& 0.17 \\
         NDF \cite{simeonov2021neural} &  0.96 & 0.92 & 0.88  & \textbf{0.78} &0.75 & 0.58\\
         \textbf{TAX-Pose} & \textbf{0.99}& \textbf{0.97}& \textbf{0.96} &  0.75& \textbf{0.84}& \textbf{0.63}\\
         
    \hline
        %  & \multicolumn{3}{c}{\textit{Avg Pairwise Angular Error $< 5^{\circ}$}}\\

    % \textbf{Arbitrary Pose} & & & \\
    %  \hline
    %      DON \cite{florence2018dense} & 0.35 & 0.45& 0.17 \\
    %      NDF \cite{simeonov2021neural} & \textbf{0.78} &0.75 & 0.58\\
    %       TAX-Pose (Ours)  &  0.75& \textbf{0.84}& \textbf{0.63}\\
         
    % \hline
    \end{tabular}
    \label{tab:mug_on_rack_exp2}
    %\end{table}
    }
    \captionof{table}{Mug on rack simulation success rate ($\uparrow$)}
    
    \end{minipage}%
    \hfill
    \noindent \begin{minipage}[t]{.24\textwidth}
    \label{tab:mug_on_rack_few_shot}
    % \small
        \centering
        \resizebox{\textwidth}{!}{
        \begin{tabular}{c|c c c}
            \hline
            \textbf{Model} & \multicolumn{3}{c} \#\textbf{Demos Used }\\
            \hline
            & 1 & 5 & 10\\
            \hline
            DON \cite{florence2018dense}  & 0.32 &0.36 & 0.45\\
            NDF \cite{simeonov2021neural} &  0.46 & 0.70 & 0.88 \\
            \textbf{TAX-Pose} & \textbf{0.77}& \textbf{0.90}& \textbf{0.96}\\
            \hline
        \end{tabular}
        }
        \vspace{-2pt}
        
        \captionof{table}{ \# Demos vs. Overall success rate ($\uparrow$)}
    \end{minipage}%
    \hfill
    \begin{minipage}[t]{.28\textwidth}
        % \small
        \centering
        \label{tab:mug_rack_ablation}
        \resizebox{\textwidth}{!}{
        \begin{tabular}{c|c c c}
            \hline
            \textbf{Ablation} & \textbf{Grasp} & \textbf{Place} & \textbf{Overall}\\
            \hline
            No Res.  & 0.97 &0.96 & 0.93\\
            Unw. SVD &  0.92 & 0.94 & 0.88 \\
            %  3-Layer MLP In Place of Attention &  0.90 & 0.82 & 0.76 \\
            No Attn. &  0.90 & 0.82 & 0.76 \\
            \textbf{TAX-Pose} & \textbf{0.99}& \textbf{0.97}& \textbf{0.96}\\
            \hline
        \end{tabular}
        }
        \captionof{table}{Mug hanging ablations success rate ($\uparrow$)}

    \end{minipage}%
% \end{table}

% \input{tables/ablation_fewshot}

% \textbf{Ablation Analysis.} 
% \textit{Number of Demonstrations.} To study the effects of number of demonstrations used on the performance of our method, we report quantitative performance of our method alongside baseline methods trained on different numbers of demonstrations (10, 5, 1) for upright pose mug hanging task as seen in Table 4. Our method outperforms the baselines for all number of demonstrations; TAX-Pose can perform well even with just 5 demonstrations.
%~\ref{tab:mug_on_rack_few_shot}
%We find that our method is able to retain high success rates relatively better than the baselines with fewer demonstrations used.\newline
% \input{tables/mug_on_rack_few_shot}

% \textit{Cross-Pose Estimation Design Choices.} We analyze the effects of the different design choices made in our Cross-Pose estimation algorithm for the upright pose mug hanging task. Specifically, we analyze the effects of 1) computing residual correspondence; 2) the use of \textit{weighted} Procrustes over the non-weighted in computing cross-pose; 3) using a transformer as the cross-object attention as in Figure \ref{fig:gc-ipm} method, as opposed to simpler model such as a 3-layer MLP. We report results in Table 5. %~\ref{tab:mug_rack_ablation}.
% \input{tables/mug_rack_ablation}
\begin{figure}
    \centering
    \includegraphics[width=\textwidth]{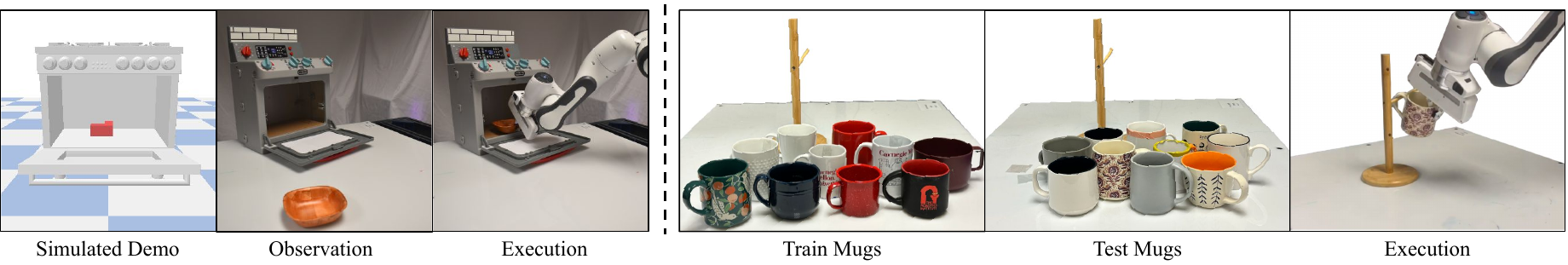}
     \vspace{-20pt}
    \caption{Real-world experiments summary. \textbf{Left: } In object placement task, we train using simulated demonstrations and test on real-world objects. \textbf{Right: }Mug Hanging real-world experiments. We train from just 10 demonstrations from 10 training mugs in the real world and test on 10 unseen test mugs. 
    %In real-world experiments, the robot observes the starting segmented point clouds of the mug and the rack, and infers the cross-pose between them using TAX-Pose. The robot then grasps the mug using a primitive and once contact is made, the robot executes a motion plan that executes the predicted relative transform. \dave{I'm not sure this whole explanation is necessary in the caption}
    }
    \label{fig:mughangingrw}
    
\end{figure}

\label{sec:mug_hanging}
\textbf{Real-World Experiments: } 
We explore the 
%extension of the 
hanging component of the mug on a rack task in a real world environment, which requires 
estimating the cross-pose between the mug and the rack.
%Our real-world mug-hanging experiments show our method's ability to make precise pose predictions on real-world data. 
We train TAX-Pose using real demonstrations of 10 different mugs hung on a rack (1 demonstration each, for a total of only 10 real-world demonstrations for training). A motion primitive is used to grasp each mug, after which the robot plans a trajectory to apply the predicted cross-pose to the grasped mug. 
%, and restrict the learning task to predicting the relationship between the mug and rack (ignoring the gripper-mug relationship). 
We evaluate the model on the 10 training mugs in novel poses, as well as on  10 unseen mugs (see Figure~\ref{fig:mughangingrw}). For each of the 20 mugs, we conduct 5 trials, varying the mug's and rack's starting poses in each trial. Success is recorded if a peg penetrates the mug handle at the end of the trial. Our method achieves a success rate of 62\% on training mugs in novel poses and 54\% on unseen mugs. A visualization of the results can be seen in Figure~\ref{fig:mughangingrw} (right) and on the \href{https://sites.google.com/view/tax-pose/home}{project website}. Note that our method is able to perform the mug hanging task while varying the pose of the mug rack (see our \href{https://sites.google.com/view/tax-pose/home}{project website}), whereas the baselines (NDF~\cite{simeonov2021neural}, DON~\cite{florence2018dense}) cannot because they assume a fixed, known rack position (see NDF~\cite{simeonov2021neural} for baseline details).

% While our method is able to accurately predict relative transforms required to achieve a given task, it does require an accurate segmentation of the objects of importance. Additionally, while our method is tested with some occlusions, it performs better with a mostly complete cloud of the object being manipulated. This means that multiple views of that object must be captured. This can be done with multiple cameras, or by lifting the object and capturing multiple views. 
% Additionally, in our experiments, we have found that the predicted rotation is not always accurate enough to hang the virtual mugs given the small number of demonstrations. In the case that more accurate estimates are required,
% Additionally, as our method is a function of correspondences, multimodal target configurations can cause potential problems. Such multimodalites can occur when interacting with symmetric objects or when the task has multiple valid configurations, such as objects with multiple valid placement surfaces, or racks with multiple usable hangers. These problems could be alleviated using a consensus-based method for mapping from multimodal soft correspondences to a single transform.  We leave this for future work.

% results suggest that the accuracy of our predicted rotations can ..\ben{Mention goal sets}

\subsection{PartNet-Mobility  Placement Tasks}
\textbf{Task Description}: We also define a PartNet-Mobility Placement task as placing a given action object relative to an anchor object based on a semantic goal position. We select a set of household furniture objects from the PartNet-Mobility dataset~\cite{Xiang2020-oz} as the anchor objects, and a set of small rigid objects released with the Ravens simulation environment \cite{zeng2020transporter, yao2023apla, jin2024multi, shen2024diffclip} as the action objects. For each anchor object, we define a set of semantic goal positions (i.e. `top', `left', `right', `in'), where action objects should be placed relative to each anchor. Each semantic goal position defines a unique task in our cross-pose prediction framework. Given a synthetic point cloud observation of both objects, the task is to predict a cross-pose that places the object at the specific semantic goal. 
We evaluate both a \emph{goal-conditioned} variant (\textbf{TAX-Pose GC}), which is trained across all goals, and a \emph{task-specific} variant (\textbf{TAX-Pose}) of our model, which trains a separate model per goal type (see Appendix~\ref{appendix:results-pm-gc} for details). In both cases we train only 1 model across all action and anchor objects.
All models are trained entirely on simulated data and transfer directly to real-world with no finetuning. Further task details can be found in the Appendix \ref{appendix:taskdetails-pm}.

\textbf{Baselines}: We compare our method to a variety of end-to-end imitation-learning-based methods trained from a motion planner expert in simulation (see Appendix \ref{appendix:taskdetails-pm-baselines} for details). Note that in the PartNet-Mobility Placement experiments, the pose of the anchor object poses are randomly varied. As such, we omit a comparison to methods that assume a static anchor, such as the Neural Descriptor Field (NDF)~\cite{simeonov2021neural, lim2021planar} and Dense Object Nets (DON)~\cite{florence2018dense, lim2022real2sim2real} baselines used in the mug hanging task (Section~\ref{sec:NDF Tasks}), as both methods
%DON~\cite{florence2018dense} and NDF~\cite{simeonov2021neural} 
%compare directly to the objects demonstration poses and 
assume that the anchor objects are in a fixed, known position. 
\begin{wraptable}[12]{r}{0.26\linewidth}
\vspace{-12pt}
% \begin{minipage}{\textwidth}
% % \begin{minipage}[b]{0.4\linewidth}
% %     \centering
% %     \includegraphics[width=\textwidth]{figures/realworld-setup.pdf}
% %     \captionof{figure}{Real-world experiments illustration. \textbf{Left: }work-space setup for physical experiments. \textbf{Center: }Octomap visualization of the perceived anchor object.} 
% %     \label{fig:image}
% % \end{minipage}%
% % % \qquad
% % \hfill
% \begin{minipage}[b]{0.5\linewidth}
%     \renewcommand{\arraystretch}{1.2}
    \resizebox{0.9\textwidth}{!}{
\begin{tabular}{lll}
\hline
\multicolumn{1}{|l|}{\multirow{2}{*}{}} & \multicolumn{2}{c|}{Average} \\
\multicolumn{1}{|l|}{} & \multicolumn{1}{c}{$\mathcal{E}_\mathbf{R}$} & \multicolumn{1}{c|}{$\mathcal{E}_\mathbf{t}$} \\ \hline
\multicolumn{1}{|l|}{E2E BC} & 42.26 & \multicolumn{1}{c|}{0.73} \\ \hline
\multicolumn{1}{|l|}{E2E DAgger} & 37.96 & \multicolumn{1}{l|}{0.69} \\ \hline
\multicolumn{1}{|l|}{Traj. Flow} & 35.95 & \multicolumn{1}{l|}{0.67} \\ \hline
\multicolumn{1}{|l|}{Goal Flow} & 26.64 & \multicolumn{1}{l|}{0.17} \\ \hline
\multicolumn{1}{|l|}{TAX-Pose} & 6.64 & \multicolumn{1}{l|}{\textbf{0.16}} \\ \hline
% \multicolumn{1}{|l|}{TAX-Pose GC} & 7.74 & \multicolumn{1}{l|}{0.17} \\ \hline
\multicolumn{1}{|l|}{\textbf{TAX-Pose GC}} & \textbf{4.94} & \multicolumn{1}{l|}{\textbf{0.16}} \\ \hline

 &  &  \\ \hline
\multicolumn{1}{|l|}{} & \multicolumn{2}{c|}{Average SR} \\ \hline
\multicolumn{1}{|l|}{Goal Flow} & \multicolumn{2}{c|}{0.31} \\ \hline
\multicolumn{1}{|l|}{\textbf{TAX-Pose}} & \multicolumn{2}{c|}{\textbf{0.92}} \\ \hline
\end{tabular}

}
% \vspace{10pt}
\caption{\textbf{Top}: Simulation Rotational ($^\circ$) and Translational ($\mathrm{m}$) Errors ($\downarrow$). \textbf{Bottom}: Real-world goal placement success rate ($\uparrow$).}
\label{tab:2tables}
\end{wraptable}

\textbf{Results}:  We report rotation ($\mathcal{E}_\mathbf{R}$) and translation ($\mathcal{E}_\mathbf{t}$) error between our predicted transform and the ground truth as geodesic rotational distance~\cite{huynh2009metrics, hartley2013rotation, elmquist2022art, teng2024gmkf} and $L2$ distance, respectively. In both our simulated experiments (Table~\ref{tab:2tables} Top) and our real-world experiments (Table~\ref{tab:2tables} Bottom), we find that TAX-Pose outperforms the baseline end-to-end imitation learning methods, with the \emph{goal-conditioned} variant, TAX-Pose GC, performing the best. 
%In simulated experiments, while direct regression via goal-flow outperforms TAX-Pose in some rare cases of translation prediction, TAX-Pose performs substantially better than all other baselines in rotation prediction. 
In real-world experiments, our method generalizes to novel distributions of starting poses better than the Goal Flow baseline, placing action objects into the goal regions with a 92\% success rate.
See Figure~\ref{fig:mughangingrw} (left) and the \href{https://sites.google.com/view/tax-pose/home}{website} for results; 
see Appendix \ref{appendix:results-pm} for more detailed tables and Appendix \ref{appendix:taskdetails-pm-baselines} for baseline details.

\section{Conclusion and Limitations}
In this paper, we show that dense soft correspondence can be used to learn task specific object relationships that generalize to novel object instances. Correspondence residuals allow our method to estimate correspondences to virtual points, outside of the objects convex hull, drastically increasing the number of tasks this method can complete. We further show that this ``cross-pose" can be learned for a task, using a small number of demonstrations. Finally, we show that our method far outperforms the baselines on two challenging tasks in both real and simulated experiments. While our method is able to predict relative pose relationships with high precision, it has several limitations:
\begin{itemize}[leftmargin=*]
    \item \textbf{Requires segmentation}: Our method requires an accurate segmentation of two objects in order to predict their relative goal pose.
    \item \textbf{Performance degrades under occlusion}: Our method performs best when complete point clouds are provided, captured via multiple cameras or by repeatedly reorienting the objects.
    \item \textbf{Poorly defined for multimodal relationships}: Because our method extracts a single global estimate of relative pose from a fixed set of correspondences, performance on objects with multiple valid goals is not well-defined. Our method might be augmented with a consensus-based or sampling-based approach to capture the multimodality of the solution space in these cases. We leave this for future work.
\end{itemize}
\section*{Acknowledgements}
{\footnotesize
This material is based upon work supported by the National Science Foundation under Grant No. IIS-1849154. This work was also supported by LG Electronics. We are grateful to Daniel Seita and Jenny Wang for their helpful feedback and discussion on the paper.
}
% \putbib[ref]
% \end{bibunit}
\bibliography{ref}  % .bib

% \begin{bibunit}[corlabbrvnat]
\appendix
\newpage
%\onecolumn

% \documentclass{article}
% \usepackage[final]{corl_2022}
%% \usepackage{corl_2022} % Use this for the initial submission.
% \usepackage{times}

% % numbers option provides compact numerical references in the text. 
% \usepackage[numbers]{natbib}
% \usepackage{caption}
% \usepackage{subcaption}
% \usepackage{wrapfig}
% \usepackage{adjustbox}
% \usepackage{floatrow, amsthm, amsthm }
% \renewcommand{\thetable}{S\arabic{table}}
% \renewcommand\thefigure{S.\arabic{figure}} 
% \newcommand{\rebuttal}[1]{{\color{magenta} #1}}
% \usepackage{color, soul} % This allows us to highlight text that you want your co-authors to review
% % \renewcommand\hl[1]{#1} % Uncomment this to turn off highlighting
% % use \hl{text} to highlight something
% % Table float box with bottom caption, box width adjusted to content
% \newcommand{\cameraready}[1]{{\color{blue} #1}}
%\newfloatcommand{capbtabbox}{table}[][\FBwidth]

%\input{math_definitions.tex}

\newtheorem{theorem}{Theorem}[section]
\newtheorem{corollary}{Corollary}[theorem]
\newtheorem{lemma}[theorem]{Lemma}
\newtheorem{definition}[theorem]{Definition}

%\beginsupplement

% \tableofcontents
\etocsetlevel{appendixplaceholder}{-1}
\etoctoccontentsline*{appendixplaceholder}{APPENDIX}{-1}

%\usepackage[font={footnotesize}]{caption}
%\input{preamble.tex}
%\begin{document}
\title{TAX-Pose: Task-Specific Cross-Pose Estimation\\ for Robot Manipulation - Appendix}
\author{}

\maketitle

%\tableofcontents
\localtableofcontents

\appendix

\section{Visual Explanations of TAX-Pose}
\label{appendix:visual_explanation}

\subsection{Illustration of Corrected Virtual Correspondence} 
\label{appendix:illustration}

\begin{figure}[H]
    \centering
    \includegraphics[width=0.7\textwidth]{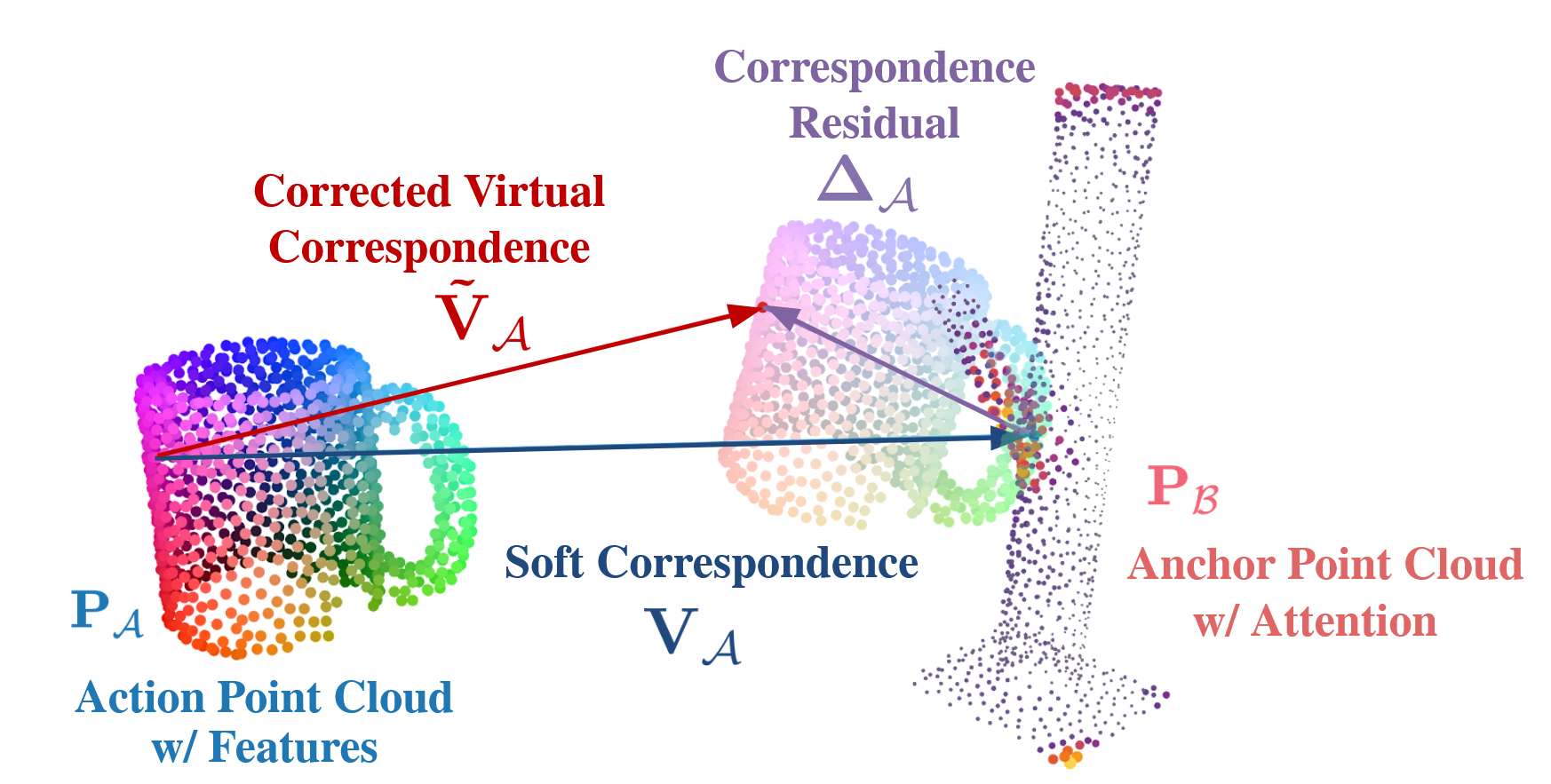}
    \caption{Computation of Corrected Virtual Correspondence. Given a pair of object point clouds $\mathbf{P}_\mathcal{A}, \mathbf{P}_\mathcal{B}$, a per-point \textit{soft correspondence} $\mathbf{V}_{\mathcal{A}}$ is first computed. Next, to allow the predicted correspondence to lie beyond object's convex hull, these soft correspondences are adjusted with \textit{correspondence residuals},  $\boldsymbol{\Delta}_{\mathcal{A}}$, which results in the \textit{corrected virtual correspondence}, $\tilde{\mathbf{V}}_{\MA}$. The coloring scheme and the point size on the rack represent the the value of the the attention weights, where the more red and larger the point, the higher the attention weights, the more gray and smaller the point the lower the attention weights. }
    \label{fig:correspondence}
\end{figure}

The virtual corresponding points, $\mathbf{V}_{\MA}$, $\mathbf{V}_{\MB}$ given by Equation 3 in the main text, are constrained to be within the convex hull of each object. However, correspondences which are constrained to the convex hull are insufficient to express a large class of desired tasks. For instance, we might want a point on the handle of a teapot to correspond to some point above a stovetop, which lies outside the convex hull of the points on the stovetop. To allow for such placements, for each point-wise embedding $\boldsymbol{\phi}_i$, we further learn a \textit{residual vector}, $\boldsymbol{\delta}_{i}^{\mathcal{A}} \in \boldsymbol{\Delta}_{\mathcal{A}}$ that corrects each virtual corresponding point, allowing us to displace each virtual corresponding point to any arbitrary location that might be suitable for the task. Concretely, we use a point-wise neural network $g_{\mathcal{R}}$ which maps each embedding into a 3D residual vector:
\begin{equation*}
    \boldsymbol{\delta}_i^{\mathcal{A}} = g_{\mathcal{R}} \left(\boldsymbol{\phi}_i^{\MA}\right) \in \mathbb{R}^3,\;\; 
     \boldsymbol{\delta}_i^{\mathcal{B}} = g_{\mathcal{R}} \left(\boldsymbol{\phi}_i^{\MB}\right) \in \mathbb{R}^3
\end{equation*}
Applying these to the virtual points, we get a set of \textit{corrected virtual correspondences}, $\tilde{\mathbf{v}}_i^{\mathcal{A}} \in \VA$ and $\tilde{\mathbf{v}}_i^{\mathcal{B}} \in \VB$,  defined as

\begin{equation}
    \tilde{\mathbf{v}}_i^{\mathcal{A}} = \mathbf{v}_i^{\mathcal{A}} + \boldsymbol{\delta}_i^{\mathcal{A}},\;\; 
    \tilde{\mathbf{v}}_i^{\mathcal{B}} = \mathbf{v}_i^{\mathcal{B}} + \boldsymbol{\delta}_i^{\mathcal{B}} 
\end{equation}
These corrected virtual correspondences $\tilde{\mathbf{v}}_i^{\mathcal{A}}$ define the estimated goal location relative to object $\MB$ for each point $\mathbf{p}_i \in \PA$ in object $\MA$, and likewise for each point in object $\MB$, as shown in Figure~\ref{fig:correspondence}.

\subsection{Learned Importance Weights}
\label{appendix:importance_weights}
\begin{figure}[H]
    \centering
    \includegraphics[width=0.7\textwidth]{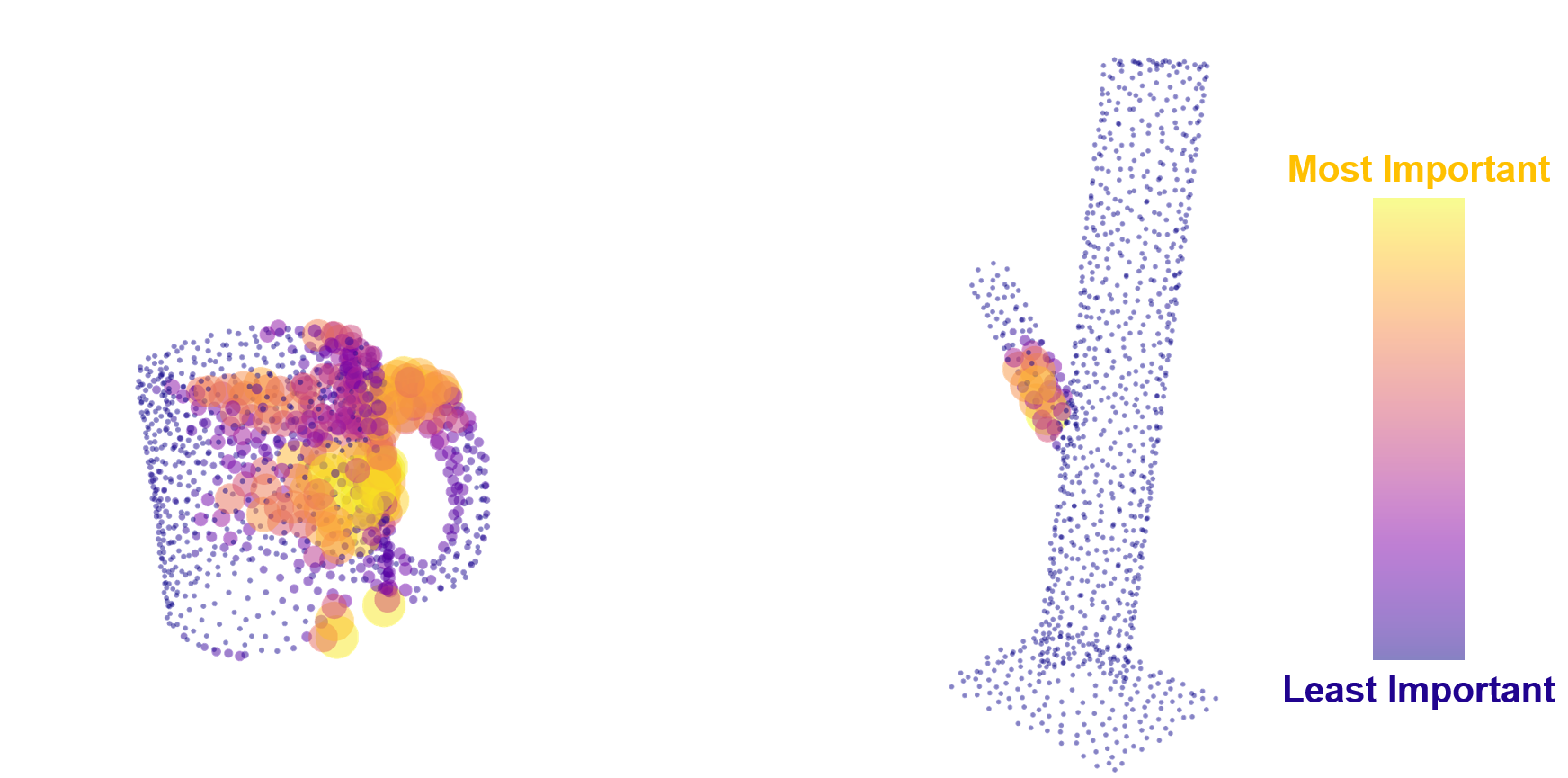}
    \caption{Learned Importance Weights for Weighted SVD on Mug and Rack.  The coloring scheme and the point size on both objects represent the the value of the the learned importance weights, where the more yellow and larger the point, the higher the learned importance weights, the more purple and smaller the point the lower the  learned importance weights. }
    \label{fig:importance_weights}
\end{figure}

A visualization of the learned importance weights, $\alpha_\mathbf{\mathcal{A}}$ and $\alpha_\mathbf{\mathcal{B}}$ for the mug and rack are visualized by both color scheme and point size in Figure~\ref{fig:importance_weights}
\section{Proof of TAX-Pose Translational Equivariance}
\label{appendix:proof}

One benefit of our method is that it is translationally equivariant by construction. This mean that if the object point clouds, $\PA$ and $\PB$, are translated by random translation $\mathbf{t}_\alpha$ and $\mathbf{t}_\beta$, respectively, i.e. $\mathbf{P}_{\mathcal{A}'} = \PA + \mathbf{t}_\alpha$ and $\mathbf{P}_{\mathcal{B}'} = \PB + \mathbf{t}_\beta$, then the resulting corrected virtual correspondences, $\VB$ and $\VA$, respectively, are transformed accordingly, i.e. $\VB + \mathbf{t}_\beta$ and $\VA + \mathbf{t}_\alpha$, respectively, as we will show below. This results in an estimated cross-pose transformation that is also equivariant to translation by construction.
This is achieved because our learned features and correspondence residuals are invariant to translation, and our virtual correspondence points are equivariant to translation.

First, our point features are a function of centered point clouds. That is, given point clouds $\mathbf{P}_\mathcal{A}$ and $\mathbf{P}_\mathcal{B}$, the mean of each point cloud is computed as 
\begin{equation*}
    \bar{\mathbf{p}}_k = \frac{1}{N_k} \sum_{i=1}^{N_k} \mathbf{P}_k.
\end{equation*}
This mean is then subtracted from the clouds,
\begin{equation*}
\mathbf{\Bar{P}}_k = \mathbf{P}_k - \bar{\mathbf{p}}_k,
\end{equation*}
which centers  the cloud at the origin.  The features are then computed on the centered point clouds:
\begin{equation*}
\mathbf{\Phi}_k = g_k(\mathbf{\Bar{P}}_k).
\end{equation*}

Since the point clouds are centered before features are computed, the features $\mathbf{\Phi}_k$ are invariant to an arbitrary translation $\mathbf{P}_{k'} = \mathbf{P}_k + \mathbf{t}_\kappa$. 

These translationally invariant features are then used, along with the original point clouds, to compute ``corrected virtual points" as a combination of virtual correspondence points, $\mathbf{v}_i^{k'}$ and the correspondence residuals, $\boldsymbol{\delta}_i^{k'}$. 
As we will see below, the ``corrected virtual points" will be translationally equivariant by construction.

The virtual correspondence points, $\mathbf{v}_i^{k'}$, are computed using weights that are a function of only the translationally invariant query and key values from the cross-object attention transformer $g_{\mathcal{T}_{\mathcal{K}}}$, $\mathbf{Q}_{\mathcal{K}}$ and $\mathbf{K}_{\mathcal{K}}$, which are in turn functions of only the translationally invariant features, $\mathbf{\Phi}_{k}$: 
\begin{equation*}
    \mathbf{w}_{i}^{\mathcal{A}' \to \mathcal{B}'} = \text{softmax}\left(\frac{\mathbf{K_{\MB'}}\mathbf{q}_{i}^{\MA'}}{\sqrt{d}}\right)= \text{softmax}\left(\frac{\mathbf{K_{\MB}}\mathbf{q}_{i}^{\MA}}{\sqrt{d}}\right)=\mathbf{w}_{i}^{\AtoB} 
\end{equation*}
thus the weights are also translationally invariant.
These translationally invariant weights are applied to the translated cloud

\begin{equation*}
\mathbf{v}_{i}^{\mathcal{A}'} = \mathbf{P}_{\MB'} \mathbf{w}_{i}^{\AtoB} 
= (\PB +\mathbf{t}_\beta) \mathbf{w}_{i}^{\AtoB} 
= \sum_j \mathbf{p}_{j}^{\MB} \cdot w_{i,j}^{\AtoB} + \mathbf{t}_\beta \sum_j w_{i,j}^{\AtoB} = \PB \mathbf{w}_{i}^{\AtoB} + \mathbf{t}_\beta,
\end{equation*}

since $\sum_{j=1}^{N_{\MB}} w_{ij}^{\AtoB} = 1$. 
Thus the virtual correspondence points $\mathbf{v}_{i}^{\mathcal{A}'}$ are equivalently translated. 
The same logic follows for the virtual correspondence points $\mathbf{v}_{i}^{\mathcal{B}'}$. This gives us a set of translationally  equivaraint virtual correspondence points $\mathbf{v}_{i}^{\mathcal{A}'}$ and $\mathbf{v}_{i}^{\mathcal{B}'}$.

The correspondence residuals, $\boldsymbol{\delta}_i^{k'}$, are a direct function of only the translationally invariant features $\mathbf{\Phi}_k$,
\begin{equation*}
\boldsymbol{\delta}_i^{k'} = g_{\mathcal{R}_{\mathcal{K}}}(\boldsymbol{\phi}_i^{k'}) = g_{\mathcal{R}_{\mathcal{K}}}(\boldsymbol{\phi}_i^k) = \boldsymbol{\delta}_i^{k}, 
\end{equation*}
therefore they are also translationally invariant. 

Since the virtual correspondence points are translationally equivariant, $\mathbf{v}_i^{\mathcal{A}'} = \mathbf{v}_i^{\mathcal{A}} + \mathbf{t}_\beta$ and the correspondence residuals are translationally invariant, $\boldsymbol{\delta}_i^{k'} = \boldsymbol{\delta}_i^k$, the final corrected virtual correspondence points, $\tilde{\mathbf{v}}_i^{\mathcal{A}'}$, are translationally equivariant, i.e. $\tilde{\mathbf{v}}_i^{\mathcal{A}'} = \mathbf{v}_{i}^{\mathcal{A}} + \boldsymbol{\delta}_i^k + \mathbf{t}_\beta$. 
This also holds for $\tilde{\mathbf{v}}_i^{\mathcal{B}'}$, giving us the final translationally equivariant correspondences between the translated object clouds as $\left(\PA + \mathbf{t}_\alpha, \VB + \mathbf{t}_\beta \right)$ and $\left(\PB + \mathbf{t}_\beta, \VA + \mathbf{t}_\alpha \right)$, where $\VB = \begin{bmatrix} \tilde{\mathbf{v}}_1^{\MA} \dots \tilde{\mathbf{v}}_{\NA}^{\MA} \end{bmatrix}^\top$. 

As a result, the final computed transformation will be automatically adjusted accordingly. Given that we use weighted SVD to compute the optimal transform, $\mathbf{T}_{\MA \MB}$, with rotational component $\mathbf{R}_{\MA \MB}$ and translational component $\mathbf{t}_{\MA \MB}$, the optimal rotation remains unchanged if the point cloud is translated, $\mathbf{R}_{\MA' \MB'} = \mathbf{R}_{\MA \MB}$, since the rotation is computed as a function of the centered point clouds. The optimal translation is defined as 
\begin{equation*}
\mathbf{t}_{\MA \MB} := \bar{\tilde{\mathbf{v}}}_\mathcal{\mathcal{A}} - \mathbf{R}_{\MA \MB} \cdot \bar{\mathbf{p}}_{\MA},
\end{equation*}
where $\bar{\tilde{\mathbf{v}}}_\mathcal{\mathcal{A}}$ and $\bar{\mathbf{p}}_{\MA}$ are the means of the corrected virtual correspondence points, $\VB$, and the object cloud $\PA$, respectively, for object $\MA$. Therefore, the optimal translation between the translated point cloud $\mathbf{P}_\mathcal{A'}$ and corrected virtual correspondence points $\mathbf{\tilde{V}}^{\mathcal{A}'}$ is 

\begin{align*}
\mathbf{t}_{\MA' \MB'} &= \bar{\tilde{\mathbf{v}}}_\mathcal{\mathcal{A}'} - \mathbf{R}_{\MA \MB} \cdot \bar{\mathbf{p}}_{\MA'} \\
&= \bar{\tilde{\mathbf{v}}}_\mathcal{\mathcal{A}} + \mathbf{t}_\beta - \mathbf{R}_{\MA \MB} \cdot (\bar{\mathbf{p}}_{\MA} + \mathbf{t}_\alpha) \\
&= \bar{\tilde{\mathbf{v}}}_\mathcal{\mathcal{A}} + \mathbf{t}_\beta - \mathbf{R}_{\MA \MB} \cdot \bar{\mathbf{p}}_{\MA} - \mathbf{R}_{\MA \MB} \cdot  \mathbf{t}_\alpha \\
&= \mathbf{t}_{\MA \MB} + \mathbf{t}_\beta - \mathbf{R}_{\MA \MB} \cdot \mathbf{t}_\alpha
\end{align*}
 
To simplify the analysis, if we assume that, for a given example, $\mathbf{R}_{\MA \MB} = \mathbf{I}$, then we get $\mathbf{t}_{\MA' \MB'} = \mathbf{t}_{\MA \MB} + \mathbf{t}_\beta - \mathbf{t}_\alpha$, demonstrating that the computed transformation is translation-equivariant by construction.

\section{Description of Cross-Object Attention Weight Computation}
\label{appendix:cross-object-weight}

\begin{figure}
    \centering
    \includegraphics[width=\textwidth]{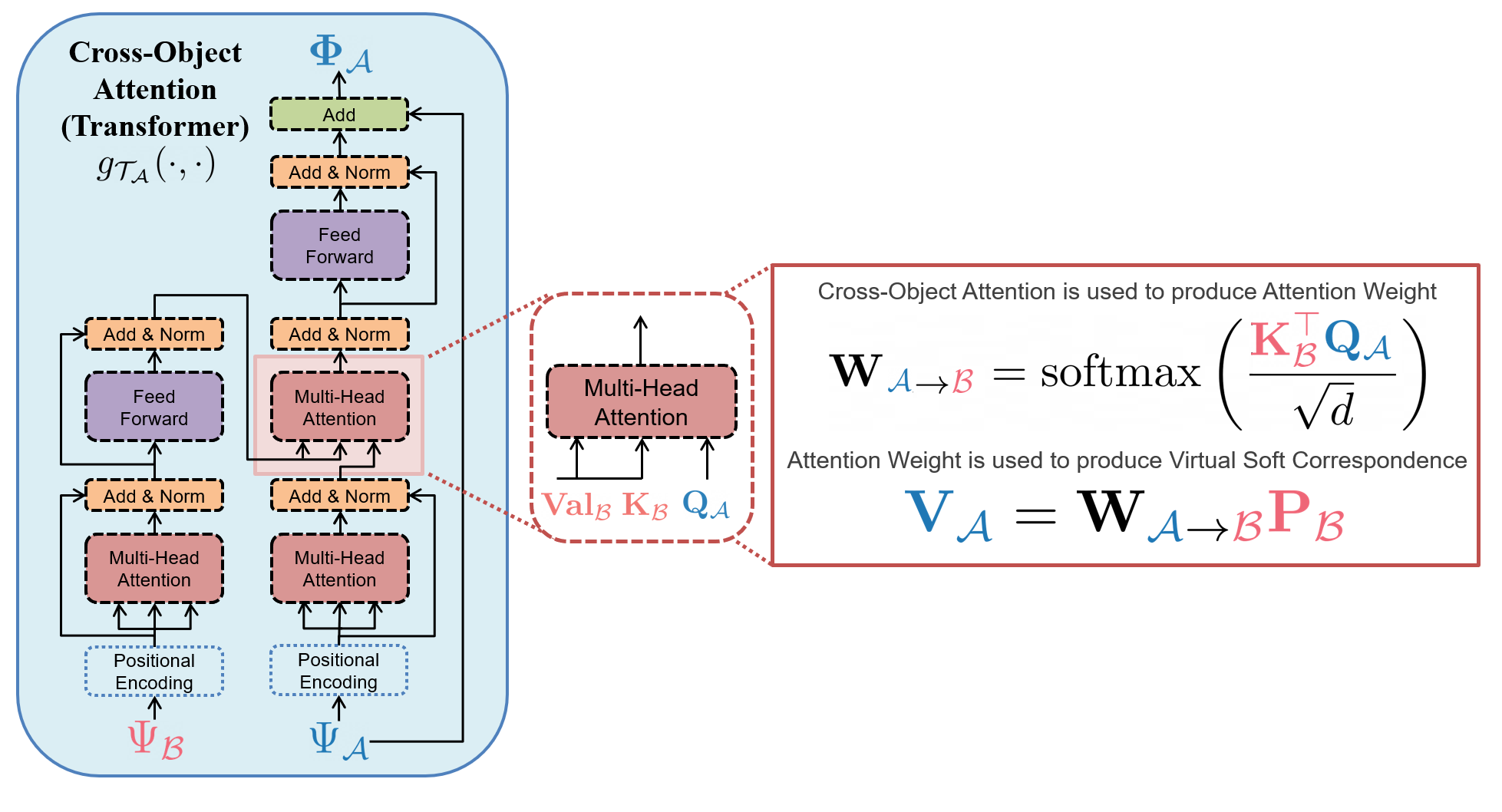}
    \caption{Cross-Object attention weight computation for virtual soft correspondence $\mathbf{V}_{\MA}$ from object $\mathcal{A}$ to $\mathcal{B}$. $\mathbf{Q_{\mathcal{K}}}, \mathbf{K_{\mathcal{K}}}, \textbf{Val}_{\mathcal{K}} \in \mathbb{R}^{\mathbf{N}_{\mathcal{K}}\times d}$ are the query, key and value (respectively) for object $\mathcal{K}$ associated with cross-object attention Transformer module $g_{\mathcal{T}_{\mathcal{K}}}$. The Transformer block is modified from Figure 2(b) in DCP~\cite{wang2019deep}.}
    \label{fig:cross-object-attention}
\end{figure}

To map our estimated features $\boldsymbol{\Psi}_{\mathcal{A}}$ and  $\boldsymbol{\Psi}_{\mathcal{B}}$ obtained from object-specific embedding networks (DGCNN), $g_{\mathcal{A}}$ and $g_{\mathcal{B}}$ respectively, to a set of normalized weight vectors $\mathbf{W}_{\AtoB}$ and $\mathbf{W}_{\BtoA}$,
% , which we call \textit{cross-object attention weights}, 
we use the cross attention mechanism of our cross-object attention Transformer module~\cite{vaswani2017attention}. 
 Following Equations 5a and 5b from the paper, we can extract the desired normalized weight vector $\mathbf{w}_{i}^{\AtoB}$ for the virtual corresponding point $\mathbf{v}_i^{\MA}$ assigned to any point $\mathbf{p}_i^{\MA}\in \mathbf{P}^{\MA}$ using the intermediate attention embeddings of cross-object attention module as:
\begin{equation}
    \mathbf{w}_{i}^{\AtoB} = \text{softmax}\left(\frac{\mathbf{K_{\MB}}\mathbf{q}_{i}^{\MA}}{\sqrt{d}}\right), \;\;
     \mathbf{w}_{i}^{\BtoA} = \text{softmax}\left(\frac{\mathbf{K_{\MA}}\mathbf{q}_{i}^{\MB}}{\sqrt{d}}\right) \;\;
     \label{eq:TAXPose-weight-computation}
\end{equation}

where $\mathbf{q}_i^{\mathcal{K}} \in \mathbf{Q_{\mathcal{K}}}$, and $\mathbf{Q_{\mathcal{K}}}, \mathbf{K_{\mathcal{K}}} \in \mathbb{R}^{\mathbf{N}_{\mathcal{K}}\times d}$ are the query and key (respectively) for object $\mathcal{K}$ associated with cross-object attention Transformer module $g_{\mathcal{T}_{\mathcal{K}}}$, as shown in Figure~\ref{fig:cross-object-attention}. These weights are then used to compute the virtual corresponding points $\mathbf{V}_{\mathcal{A}}$, $\mathbf{V}_{\mathcal{B}}$ using Equations 5a and 5b in the main paper.

\subsection{Ablation}
\label{appendix:cross-object-weight-ablation}

To explore the importance of this weight computation design choice described in Equation~\ref{eq:TAXPose-weight-computation}, we conducted an ablation experiment on this design choice against an alternative, arguably simpler method for cross-object attention weight computation that was used in prior work~\cite{wang2019deep}.
%, by computing the dot-product between point embeddings of the two objects, which we refer to as \textit{point embedding dot-product}, described as follows.
Since the point embeddings $\boldsymbol{\phi}_i^{\mathcal{A}}$ and $\boldsymbol{\phi}_i^{\mathcal{B}}$ have the same dimension $d$, we can select the inner product of the space as a similarity metric between two embeddings. 

To compute the virtual corresponding point $\mathbf{v}_i^{\MA}$ assigned to any point $\mathbf{p}_i^{\MA}\in \mathbf{P}^{\MA}$, we can extract the desired normalized weight vector $\mathbf{w}_{i}^{\AtoB}$ with the softmax function: 

\begin{equation}
    \mathbf{w}_{i}^{\AtoB} = \text{softmax}\left(\boldsymbol{\Phi}_{\MB}^\top \boldsymbol{\phi}_i^{\MA}\right), \;\;
     \mathbf{w}_{i}^{\BtoA} = \text{softmax}\left(\boldsymbol{\Phi}_{\MA}^\top \boldsymbol{\phi}_i^{\MB}\right)
     \label{eq:point embedding dot-product}
\end{equation}
This is the approach used in the prior work of Deep Closest Point (DCP)~\cite{wang2019deep}. In the experiments below, we refer to this approach as \textit{point embedding dot-product}.

We conducted an ablation experiment on the weight computation method used in TAX-Pose (Equation~\ref{eq:TAXPose-weight-computation})
against the simpler approach from DCP~\cite{wang2019deep} (Equation~\ref{eq:point embedding dot-product}),
on the upright mug hanging task in simulation.  The models are trained from 10 demonstrations and tested on 100 trials over the test mug set. As seen in Table~\ref{tab:ablation_weight}, the TAX-Pose approach (Equation~\ref{eq:TAXPose-weight-computation}) outperforms \textit{point embedding dot-product} (Equation~\ref{eq:point embedding dot-product}) in all three evaluation categories on \textit{grasp}, \textit{place}, and \textit{overall} in terms of test success rate.

\begin{table}[H]
 
    \centering
\renewcommand{\arraystretch}{1}
 
    \begin{tabular}{c|c c c}
    
        \hline
        \textbf{Attention Weight Ablation} & \textbf{Grasp} & \textbf{Place} & \textbf{Overall}\\
        \hline
        Point Embedding Dot-Product (Eqn.~\ref{eq:point embedding dot-product}) & 0.83 & 0.92 & 0.92\\
        \textbf{TAX-Pose (Ours) (Eqn.~\ref{eq:TAXPose-weight-computation})} & \textbf{0.99}& \textbf{0.97}& \textbf{0.96}\\
        \hline
    
    \end{tabular}

    \caption{Test success rate ($\uparrow$) over 100 trials for mug hanging upright task, ablated on attention weight computation methods.}
    \label{tab:ablation_weight}
\end{table}

\section{Description of Weighted SVD}
\label{appendix:weighted-svd}

The objective function for computing the optimal rotation and translation given a set of correspondences for object $\mathcal{K}$, $\{\mathbf{p}_i^k \rightarrow \tilde{\mathbf{v}}_i^k\}_i^{N_k}$ and weights $\{\alpha_i^k\}_i^{N_k}$, is as follows:
\begin{equation*}
    \mathcal{J}(\TAB) = \sum_{i=1}^{\NA} \alpha_i^{\MA} || \TAB~\mathbf{p}_{i}^{\MA} - \tilde{\mathbf{v}}_i^{\MA} ||_2^2 + \sum_{i=1}^{\NB} \alpha_i^{\MB} || \TAB^{-1}~\mathbf{p}_{i}^{\MB} - \tilde{\mathbf{v}}_i^{\MB} ||_2^2
\end{equation*}

First we center (denoted with $*$) the point clouds and virtual points independently, with respect to the learned weights, and stack them into frame-specific matrices (along with weights) retaining their relative position and correspondence:
\begin{equation*}
    \mathbf{A} = \begin{bmatrix} \PA^{*\top} & \VB^{*\top} \end{bmatrix}, \;\;
    \mathbf{B} = \begin{bmatrix} \VA^{*\top} & \PB^{*\top} \end{bmatrix}^\top, \;\;
  \boldsymbol\Gamma = \text{diag} \left(\begin{bmatrix}  \boldsymbol{\alpha}_{\MA} & \boldsymbol{\alpha}_{\MB} \end{bmatrix}\right)
\end{equation*}

Then the minimizing rotation $\mathbf{R}_{\MA \MB}$ is given by:

\begin{subequations}
\begin{minipage}{.5\textwidth}
    \begin{equation} 
        \mathbf{U}\mathbf{\Sigma}\mathbf{V}^\top = \mathrm{svd}(\mathbf{A} \boldsymbol\Gamma\mathbf{B}^\top)
    \end{equation}
\end{minipage}
\begin{minipage}{.5\textwidth}
    \begin{equation} 
        \mathbf{R}_{\MA \MB} = \mathbf{U}\mathbf{\Sigma}_*\mathbf{V}^\top
    \end{equation}
\end{minipage}
\label{SVD-full}
\end{subequations}

where $\mathbf{\Sigma}_* = \mathrm{diag}(\begin{bmatrix} 1,1, ... \mathrm{det}(\mathbf{U}\mathbf{V}^\top)\end{bmatrix}$ and $\mathrm{svd}$ is a differentiable SVD operation~\cite{papadopoulo2000estimating}.

The optimal translation can be computed as:

\begin{subequations}
\begin{minipage}{.3\textwidth}
    \begin{equation*} 
        \mathbf{t}_{\MA} = \mathbf{\bar{\tilde{v}}}_{\MB} - \mathbf{R}_{\MA \MB} \mathbf{\bar{p}}_{\MA}
    \end{equation*}
\end{minipage}%
\begin{minipage}{.3\textwidth}
    \begin{equation*} 
        \mathbf{t}_{\MB} = \mathbf{\bar{p}}_{\MB} - \mathbf{R}_{\MA \MB} \mathbf{\bar{\tilde{v}}}_{\MA}
    \end{equation*}
\end{minipage}
\begin{minipage}{.4\textwidth}
    \begin{equation} 
        \mathbf{t} = \frac{\NA}{N} \textbf{t}_{\MA} + \frac{\NB}{N} \textbf{t}_{\MB}
    \end{equation}
\end{minipage}
\label{SVD-trans}
\end{subequations}

with $N = \NA + \NB$. In the special translation-only case, the optimal translation and be computed by setting $\mathbf{R}_{\MA \MB}$ to identity in above equations. The final transform can be assembled:

\begin{equation}
    \TAB = \begin{bmatrix} \mathbf{R}_{\MA \MB} & \mathbf{t} \\
    0 & 1 \end{bmatrix}
\end{equation}
\section{Training Details}
\label{appendix:training-details}

\subsection{Supervision}
\label{appendix:training-supervision}

% When learning dense representation of a pose, there are several options for supervising the representation. 
To train the encoders $g_{\mathcal{A}}(\bar{\mathbf{P}}_\mathcal{A})$, $g_{\mathcal{B}}(\bar{\mathbf{P}}_\mathcal{B})$  as well as the residual networks $g_{\mathcal{R}_{\mathcal{A}}} \left(\boldsymbol{\phi}_i^{\MA}\right)$, $g_{\mathcal{R}_{\mathcal{B}}} \left(\boldsymbol{\phi}_i^{\MB}\right)$, we use a set of losses defined below. We assume we have access to a set of demonstrations of the task, in which the action and anchor objects are in the target relative pose such that $\TAB = \mathbf{I}$.

% with action and anchor points in the required relative pose. 

% \textbf{Direct Pose Regressions Loss}
% Given a ground truth transform $\mathbf{T}_\mathrm{GT}$, composed of rotation $\mathbf{R}_\mathrm{GT}$ and translation $\mathbf{t}_\mathrm{GT}$, we can directly supervise the the predicted pose. We use the Frobenius norm for the rotation and MSE for the translation, 
% % We supervise the predicted rotation and translation directly with the ground-truth pose.
% \begin{equation}
%     \mathcal{L}_\mathrm{pose} = \left\|\mathbf{R}^\top\mathbf{R}_\mathrm{GT} - \mathbf{I}\right\|_F^2 + \left\|\mathbf{t} - \mathbf{t}_\mathrm{GT}\right\|^2,
%     \label{equ:pose_loss}
% \end{equation}

\textbf{Point Displacement Loss~\cite{xiang2017posecnn,li2018deepim}:}
Instead of directly supervising the rotation and translation (as is done in DCP), we supervise the predicted transformation using its effect on the points. For this loss, we take the point clouds of the objects in the demonstration configuration, and transform each cloud by a random transform, $\mathbf{\hat{P}}_{\mathcal{A}} = \mathbf{T}_{\alpha} \mathbf{P}_{\mathcal{A}}$, and $\mathbf{\hat{P}}_{\mathcal{B}} = \mathbf{T}_{\beta} \mathbf{P}_{\mathcal{B}}$. This would give us a ground truth transform of $\TAB^{GT} =\mathbf{T}_{\beta}\mathbf{T}_{\alpha}^{-1}$;
%to transform object $\mathcal{A}$ to match the newly transformed object $\mathcal{B}$. 
the inverse of this transform would move object $\mathcal{B}$ to the correct position relative to object $\mathcal{A}$. Using this ground truth transform, we compute the MSE loss between the correctly transformed points and the points transformed using our prediction. 
\begin{equation}
    \mathcal{L}_\mathrm{disp} = \left\|\mathbf{T}_{\mathcal{A}\mathcal{B}} \mathbf{P}_{\mathcal{A}}  - \mathbf{T}_{\mathcal{A}\mathcal{B}}^{GT} \mathbf{P}_{\mathcal{A}}\right\|^2 + \left\|\mathbf{T}_{\mathcal{A}\mathcal{B}}^{-1} \mathbf{P}_{\mathcal{B}}  - \mathbf{T}_{\mathcal{A}\mathcal{B}}^{ GT-1} \mathbf{P}_{\mathcal{B}}\right\|^2
    \label{equ:disp_loss}
\end{equation}
\textbf{Direct Correspondence Loss.}
While the Point Displacement Loss best describes errors seen at inference time, it can lead to correspondences that are inaccurate but whose errors average to the correct pose. To improve these errors we directly supervise the learned correspondences $\tilde{V}_{\mathcal{A}}$ and $\tilde{V}_{\mathcal{B}}$: %using the below loss,
\begin{equation}
    \mathcal{L}_\mathrm{corr} = \left\| \VA - \mathbf{T}_{\mathcal{A}\mathcal{B}}^{GT} \mathbf{P}_{\mathcal{A}}\right\|^2 + \left\|\VB - \mathbf{T}_{\mathcal{A}\mathcal{B}}^{GT-1} \mathbf{P}_{\mathcal{B}}\right\|^2.
    \label{equ:corr_loss}
\end{equation}
\textbf{Correspondence Consistency Loss.}
%For a less rigid loss, that also improve the accuracy of the correspondences, 
Furthermore, a consistency loss can be used. This loss penalizes correspondences that deviate from the final predicted transform. A benefit of this loss is that it can help the network learn to respect the rigidity of the object, while it is still learning to accurately place the object. Note, that this is similar to the Direct Correspondence Loss, but uses the predicted transform as opposed to the ground truth one. As such, this loss requires no ground truth:
\begin{equation}
    \mathcal{L}_\mathrm{cons} =  \left\| \VA - \mathbf{T}_{\mathcal{A}\mathcal{B}} \mathbf{P}_{\mathcal{A}}\right\|^2 + \left\|\VB - \mathbf{T}_{\mathcal{A}\mathcal{B}}^{-1} \mathbf{P}_{\mathcal{B}}\right\|^2.
    \label{equ:cons_loss}
\end{equation}

\textbf{Overall Training Procedure. }We train with a combined loss $\mathcal{L}_\mathrm{net} = \mathcal{L}_\mathrm{disp} + 
\lambda_1 \mathcal{L}_\mathrm{corr} + 
\lambda_2 \mathcal{L}_\mathrm{cons}$,
where $\lambda_1$ and $\lambda_2$ are hyperparameters. We use a similar network architecture as DCP~\cite{wang2019deep}, which consists of DGCNN \cite{wang2019dynamic} and a Transformer \cite{vaswani2017attention}.

 In order to quickly adapt to new tasks, we optionally pre-train the DGCNN embedding networks over a large set of individual objects  %pairs of Shapenet Object point cloud and their randomly transformed counterparts 
using the InfoNCE loss \cite{oord2018representation} with a geometric distance weighting and random transformations, to learn $SE(3)$ invariant embeddings, see Appendix E.2 for details.

\subsection{Pretraining}
\label{appendix:training-pretraining}

We utilize pretraining for the embedding network for the mug hanging task, and describe the details below. 

We pretrain embedding network for each object category (mug, rack, gripper), such that the embedding network is \textit{$SE(3)$ invariant} with respect to the point clouds of that specific object category. Specifically, the mug-specific embedding network is pretrained on 200 ShapeNet \cite{chang2015shapenet} mug instances, while the rack-specific and gripper-specific embedding network is trained on the same rack and Franka gripper used at test time, respectively. Note that before our pretraining, the network is randomly initialized with the Kaiming initialization scheme~\cite{he2015delving}; we don't adopt any third-party pretrained models.

For the network to be trained to be $SE(3)$ invariant, we pre-train with InfoNCE loss \cite{oord2018representation} with a geometric distance weighting and random $SE(3)$ transformations. Specifically, given a point cloud of an object instance, $\PA$,  of a specific object category $\mathcal{A}$, and an embedding network $g_{\MA}$, we define the point-wise embedding for $\PA$ as $\Phi_{\mathcal{A}} = g_{\MA}(\PA)$, where $\phi_i^{\mathcal{A}} \in \Phi_{\mathcal{A}} $ is a $d$-dimensional vector for each point $p_{i}^{\MA} \in \PA$. Given a random $SE(3)$ transformation, $\mathbf{T}$, we define $\Psi_{\mathcal{A}}= g_{\MA}(\mathbf{T}\PA)$, where $\psi_i^{\mathcal{A}} \in \Psi_{\mathcal{A}} $ is the $d$-dimensional vector for the $i$th point $p_{i}^{\MA} \in \PA$.

The weighted contrastive loss used for pretraining,  $\mathcal{L}_{wc}$, is defined as
\begin{align}
\mathcal{L}_{wc} :&= - \sum_{i}{\log \left[\frac{\exp \left(\phi_i^\top \psi_i\right)}{\sum_{j} \exp\left({d_{ij} \left(\phi_i^\top \psi_{j}\right)}\right)}\right]} \\
    d_{ij}:&= \begin{cases}
       \frac{1}{\mu} \tanh{(\lambda \|p_i^{\mathcal{A}} - p_j^{\mathcal{A}}\|_2)},& \text{if } i \neq j\\
        1,              & \text{otherwise}
    \end{cases} \\
    \mu:&=\max{(\tanh{(\lambda \|p_i^{\mathcal{A}} - p_j^{\mathcal{A}}\|_2)})}
\end{align}
For this pretraining, we use $\lambda:=10$.

\subsection{Architectural Variants}
\label{appendix:training-architectures}

\textbf{Goal-Conditioned TAX-Pose}:  To enable a single TAX-Pose model to scale to multiple related placement sub-tasks for a pair of action and anchor objects, we design a \textbf{goal-conditioned} variant (\textbf{TAX-Pose GC}), which receives a one-hot encoding of the desired semantic goal position (e.g. `top', `left', \dots) for the task. This contextual encoding is incorporated into each DGCNN module in the same way as proposed in the original DGCNN paper. This encoding can be used to provide an embedding of the specific placement relationship that is desired in a scene (e.g. selecting a ``top'' vs. ``left'' placement position) and thus enable goal conditioned placement. 

\textbf{Vector Neurons}: We briefly experimented with Vector Neurons \cite{deng2021vector} and found that this led to worse performance on this task.

\section{Additional Results}
\label{appendix:results}

\subsection{NDF Placement Tasks}
\label{appendix:results-ndf}

\subsubsection{Further Ablations on Mug Hanging Task}
\label{appendix:results-ndf-ablations}

In order to examine the effects of different design choices in the training pipeline, we conduct ablation experiments with final task-success (\textit{grasp}, \textit{place}, \textit{overall
}) as evaluation metrics for Mug Hanging task with upright pose initialization for the following components of our method, see Table ~\ref{tab:mug_rack_ablation_full} for full ablation results along six ablated dimensions as detailed below. We also performed an ablation experiment on alternative cross-object attention weight computation, as explained in Appendix~\ref{appendix:cross-object-weight} and results can be found in Table~\ref{tab:ablation_weight}. For consistency, all ablated models are trained for 15K steps. 

\begin{enumerate}
    \item \textbf{Loss.} In the full pipeline reported, we use a weighted sum of the three types of losses described in \textbf{Section 4.2} of the paper. Specifically, the loss used $\mathcal{L}_\mathrm{net}$ is given by
    \begin{equation}
    \mathcal{L}_\mathrm{net} = \mathcal{L}_\mathrm{disp} + 
    \lambda_1 \mathcal{L}_\mathrm{cons} + \lambda_2
    \mathcal{L}_\mathrm{corr}
    \end{equation} 
    where we chose $\lambda_1 = 0.1$, $\lambda_2 = 1$ after hyperparameter search.

    We ablate usage of all three types of losses, by reporting the final task performance in simulation for all experiments, specifically, we report task success on the following $\mathcal{L}_\mathrm{net}$  variants. 
    \begin{enumerate}[nosep]
       \item Remove the point displacement loss term, $\mathcal{L}_\mathrm{disp}$, after which we are left with $$\mathcal{L'}_\mathrm{net} = (0.1) \mathcal{L}_\mathrm{cons} + 
    \mathcal{L}_\mathrm{corr}$$
        \item Remove the direct correspondence loss term, $\mathcal{L}_\mathrm{corr}$, after which we are left with $$\mathcal{L'}_\mathrm{net} =  \mathcal{L}_\mathrm{disp} + 
    (0.1)\mathcal{L}_\mathrm{cons}$$
        \item Remove the correspondence consistency loss term,  $\mathcal{L}_\mathrm{cons}$, after which we are left with $$\mathcal{L'}_\mathrm{net} =  \mathcal{L}_\mathrm{disp} + 
    \mathcal{L}_\mathrm{corr}$$
        \item From testing loss variants above, we found that the point displacement loss is a vital contributing factor for task success, where removing this loss term results in no overall task success, as shown in Table ~\ref{tab:mug_rack_ablation_full}. However, in practice, we have found that adding the correspondence consistency loss and direct correspondence loss generally help to lower the rotational error of predicted placement pose compared to the ground truth of collected demos. To further investigate the effects of the combination of these two loss terms, we used a scaled weighted combination of $\mathcal{L}_\mathrm{cons}$ and $\mathcal{L}_\mathrm{corr}$, such that the former weight of the displacement loss term is transferred to consistency loss term, with the new $\lambda_1 = 1.1$, and with $\lambda_2 = 1$ stays unchanged. Note that this is different from variant (a) above, as now the consistency loss given a comparable weight with dense correspondence loss term, which intuitively makes sense as the consistency loss is a function of the predicted transform $\mathbf{T}_{\mathcal{A}\mathcal{B}}$ to be used, while the dense correspondence loss is instead a function of the ground truth transform, $\mathbf{T}_{\mathcal{A}\mathcal{B}}^{GT}$, which provides a less direct supervision on the predicted transforms. Thus we are left with $$\mathcal{L'}_\mathrm{net} =  (1.1) \mathcal{L}_\mathrm{cons} + 
     \mathcal{L}_\mathrm{corr}$$
    \end{enumerate}
  
    \item \textbf{Usage of Correspondence Residuals.} After predicting a per-point soft correspondence between objects $\mathcal{A}$ and $\mathcal{B}$, we adjust the location of the predicted corresponding points by further predicting a point-wise correspondence residual vector to displace each of the predicted corresponding point. This allows the predicted corresponding point to get mapped to free space outside of the convex hulls of points in object $\mathcal{A}$ and $\mathcal{B}$. This is a desirable adjustment for mug hanging task, as the desirable cross-pose usually require points on the mug handle to be placed somewhere near but not in contact with the mug rack, which can be outside of the convex hull of rack points. We ablate correspondence residuals by directly using the soft correspondence prediction to find the cross-pose transform through weighted SVD, without any correspondence adjustment via correspondence residual. 
    \item \textbf{Weighted SVD vs Non-weighted SVD.} We leverage weighted SVD as described in \textbf{Section 4.1} of the paper as we leverage predicted per-point weight to signify the importance of specific correspondence. We ablate the use of weighted SVD, and we use an un-weighted SVD, where instead of using the predicted weights, each correspondence is assign equal weights of $\frac{1}{N}$, where $N$ is the number of points in the point cloud $\mathbf{P}$ used.
    \item \textbf{Pretraining.} In our full pipeline, we pretrain the point cloud embedding network such that the embedding network is $SE(3)$ invariant. Specifically, the mug-specific embedding network is pretrained on ~200 ShapeNet mug objects, while the rack-specific and gripper specific embedding network is trained on the same rack and Franka gripper used at test time, respectively. We conduct ablation experiments where 
    \begin{enumerate}[nosep]
        \item We omit the pretraining phase of embedding network
        \item We do not finetune the embedding network during downstream training with task-specific demonstrations.
    \end{enumerate}
    Note that in practice, we find that pretraining helps speed up the downstream training by about a factor of 3, while models with or without pretraining both reach a similar final performance in terms of task success after both models converge. 
    
    \item \textbf{Usage of Transformer as Cross-object Attention Module.} In the full pipeline, we use transformer as the cross-object attention module, and we ablate this design choice by replacing the transformer architecture with a simple 3-layer MLP with ReLU activation and hidden dimension of 256, and found that this leads to worse place and grasp success.
    \item \textbf{Dimension of Embedding.} In the full pipeline, the embedding is chosen to be of dimension 512. We conduct experiment on much lower dimension of 16, and found that with dimension =16, the place success is much lower, dropped from 0.97 to 0.59. 

\end{enumerate} 
\begin{table}[H]
 
    \centering
\renewcommand{\arraystretch}{1.2}  % good vertical spacing for the long text descriptions.
\resizebox{0.85\textwidth}{!}{
    \begin{tabular}{c| c c c}
    \hline
  \textbf{Ablation Experiment} & \textbf{Grasp} & \textbf{Place} & \textbf{Overall}\\
  
    %  & \multicolumn{3}{c}{\textit{Avg Pairwise Angular Error $< 1^{\circ}$}}\\
   
        % \textbf{Upright Pose} & & &\\
         \hline
        % \textit{Loss}&&&\\
          No $\mathcal{L}_\mathrm{disp}$ &  0.01 & 0 & 0 \\
          No $ \mathcal{L}_\mathrm{corr}$ &  0.89 & 0.91 & 0.84 \\
           No $  \mathcal{L}_\mathrm{cons}$ &  \textbf{0.99} & 0.95 & 0.94 \\
            Scaled Combination: 
            %of  $ \mathcal{L}_\mathrm{corr} \&   \mathcal{L}_\mathrm{cons}$ (
            $ 1.1\mathcal{L}_\mathrm{cons} + \mathcal{L}_\mathrm{corr}$ &  0.10 & 0.01 & 0.01 \\
         \hline 
          No Adjustment via Correspondence Residuals  & 0.97 &0.96 & 0.93\\
         \hline 
         Unweighted SVD &  0.92 & 0.94 & 0.88 \\
        \hline 
         No Finetuning for Embedding Network &  0.98 & 0.93 & 0.91 \\
         No Pretraining for Embedding Network &  \textbf{0.99} & 0.72 & 0.71 \\
         \hline 
         3-Layer MLP In Place of Transformer &  0.90 & 0.82 & 0.76 \\
        \hline
        Embedding Network Feature Dim = 16 & 0.98 & 0.59 & 0.57 \\
    
        \hline
        \hline
        \textbf{TAX-Pose (Ours)}  & \textbf{0.99}& \textbf{0.97}& \textbf{0.96}\\

    \hline
    \end{tabular}
    }
    \caption{Mug Hanging Ablations Results}
    \label{tab:mug_rack_ablation_full}
\end{table}

\subsubsection{Effects of Pretraining on Mug Hanging Task}
\label{appendix:results-ndf-pretraining}

We explore the effects of pretraining on the final task performance, as well as training convergence speed. We have found that pretraining the point cloud embedding network as described in~\ref{appendix:training-pretraining}, is a helpful but not necessary component in our training pipeline. Specifically, we find that while utilizing pretraining reduces training time, allowing the model to reach similar task performance and train rotation/translation error with much fewer training steps, this component is not necessary if training time is not of concern. In fact, as see in Table~\ref{tab:no_pretraining_longer}, we find that for mug hanging tasks, by training the models from scratch without our pretraining, the models are able to reach similar level of task performance of $0.99$ grasp, $0.92$ for place and $0.92$ for overall success rate. Furthermore, it is able to achieve similar level of train rotation error of $4.91^{\circ}$ and translation error of $0.01m$, compared to the models with pretraining. However, without pre-trainig, the model needs to be trained for about 2 times longer (26K steps compared to 15K steps) to reach the similar level of performance. Thus we adopt our object-level pretraining in our overall pipeline to allow lower training time. 

Another benefit of pretraining is that the pretraining for each object category is done in a task-agnostic way, so the network can be more quickly adapted to new tasks after the pretraining is performed. For example, we use the same pre-trained mug embeddings for both the gripper-mug cross-pose estimation for grasping as well as the mug-rack cross-pose estimation for mug hanging.

\begin{table}[H]
 
    \centering
\renewcommand{\arraystretch}{1.2}
\resizebox{\textwidth}{!}{
    \begin{tabular}{c| c c c|c c}
    \hline
  \textbf{Ablation Experiment} & \textbf{Grasp} & \textbf{Place} & \textbf{Overall} & \textbf{Train Rotation Error} & \textbf{Train Translation Error}  \\
   &   &   &   & \textbf{ (${}^{\circ}$)} & \textbf{(m)}  \\
  \hline
   \hline
  
         No Pre-Training for Embedding Network &  \multirow{2}{*}{\textbf{0.99}} &\multirow{2}{*}{ 0.92} & \multirow{2}{*}{0.92} & \multirow{2}{*}{4.91} & \multirow{2}{*}{\textbf{0.01}} \\ 
         (trained for 26K steps) &   &  &  &   & \\ 
         \hline 
        No Pre-training for Embedding Network &  \multirow{2}{*}{\textbf{0.99}} &\multirow{2}{*}{0.72} & \multirow{2}{*}{0.71} & \multirow{2}{*}{15.39} & \multirow{2}{*}{\textbf{ 0.01}} \\
          (trained for 15K steps) &   &  &  &  & \\ 
     
         \hline
        \textbf{TAX-Pose (Ours)} &  \multirow{2}{*}{\textbf{0.99}} &\multirow{2}{*}{\textbf{0.97}} & \multirow{2}{*}{\textbf{0.96}} & \multirow{2}{*}{\textbf{4.33}} & \multirow{2}{*}{\textbf{ 0.01}} \\ 
         (trained for 15K steps) &   &  &  &  & \\ 
    \hline
    \end{tabular}
    }
    \caption{Ablation Experiments on the Effects of Pre-Training. We report the task success rate for upright mug hanging task over 100 trials each, as well as the grasping model's training rotational error (${}^{\circ}$) and translation error (m).}
    \label{tab:no_pretraining_longer}
\end{table}

\subsubsection{Additional Simulation Experiments on Bowl and Bottle Placement Task}
\label{appendix:results-ndf-bowlbottle}

\begin{table}[H]
    % \small
    \centering
     
\resizebox{0.7\textwidth}{!}{
    \begin{tabular}{c|c|c c c| c c c}
    \hline
   \textbf{Object} & \textbf{Algorithm}& \textbf{Grasp} & \textbf{Place} & \textbf{Overall} & \textbf{Grasp} & \textbf{Place} & \textbf{Overall}\\
    % & Grasp & Place & Overall\\
    \hline
    %  & \multicolumn{3}{c}{\textit{Avg Pairwise Angular Error $< 1^{\circ}$}}\\
    & & \multicolumn{3}{c|}{\textbf{Upright Pose}}& \multicolumn{3}{c}{\textbf{Arbitrary Pose}} \\
         \hline
    \multicolumn{1}{l|}{\multirow{3}{*}{\textbf{Mug}}} & DON \cite{florence2018dense} & 0.91 & 0.50 & 0.45 & 0.35 & 0.45& 0.17 \\
        & NDF \cite{simeonov2021neural} &  0.96 & 0.92 & 0.88  & \textbf{0.78} &0.75 & 0.58\\
        & \textbf{TAX-Pose (Ours)} & \textbf{0.99}& \textbf{0.97}& \textbf{0.96} &  0.75 & \textbf{0.84}& \textbf{0.63}\\
    \hline
    
    \multicolumn{1}{l|}{\multirow{3}{*}{\textbf{Bowl}}} & DON \cite{florence2018dense} & 0.50 & 0.35 & 0.11 & 0.08 & 0.20 & 0 \\
        & NDF \cite{simeonov2021neural} & 0.91 & \textbf{1} & 0.91  & \textbf{0.79} & \textbf{0.97} & 0.78\\
        & \textbf{TAX-Pose (Ours)} & \textbf{0.99}& 0.92 & \textbf{0.92} &  0.74 & 0.85& \textbf{0.85}\\
    \hline
    
    \multicolumn{1}{l|}{\multirow{3}{*}{\textbf{Bottle}}} & DON \cite{florence2018dense} & 0.79 & 0.24 & 0.24 & 0.05 & 0.02 & 0.01 \\
        & NDF \cite{simeonov2021neural} & \textbf{0.87} & \textbf{1} & \textbf{0.87} & \textbf{0.78} & \textbf{0.99} & \textbf{0.77} \\
        & \textbf{TAX-Pose (Ours)} & 0.55 & 0.99 & 0.55 &  0.61 & 0.55 & 0.52\\
         
    \hline
 
    \end{tabular}
    }
    \caption{Unseen Object Instance Manipulation Task Success Rates ($\uparrow$) in Simulation on \textit{Mug}, \textit{Bowl} and \textit{Bottle} for Upright and Arbitrary Initial Pose. Each result is the success rate over 100 trials.} 
    
    \label{tab:bottle_bowl}
\end{table}
 
Additional results on \textit{Grasp}, \textit{Place} and \textit{Overall} success rate in simulation for \textbf{Bowl} and \textbf{Bottle} are shown in Table~\ref{tab:bottle_bowl}. For bottle and bowl experiment, we follow the same experimentation setup as in~\cite{simeonov2021neural}, where the successful $\textit{grasp}$ is considered if a stable grasp of the object is obtained, and a successful $\textit{place}$ is considered when the bottle or bowl is stably placed upright on the elevated flat slab over the table without falling on the table. Reported task success results in are for both \textit{Upright Pose} and \textit{Arbitrary Pose} run over 100 trials each.

Unlike mugs, bowls and bottles exhibit rotational object symmetry, which we have found cause the trained model to perform poorly in the \textit{Grasp} task. To mitigate this, we applied symmetry breaking techniques for the \textbf{Bowl} and \textbf{Bottle} placement tasks by algorithmically creating symmetry labels, $l_i^{\mathcal{K}} \in [-1, 1]$ for object $\mathcal{K}$, as continuous real numbers between -1 and 1 inclusive for each point $\mathbf{p}_i^{\mathcal{K}}$ during training and testing. The symmetry label for $i$-th point is concatenated with the associated point-wise embedding, $\mathbf{\psi}_i^{\mathcal{K}}$, resulting in an augmented point-wise embedding, $\hat{\mathbf{\psi}}_i^{\mathcal{K}} := \begin{bmatrix}\mathbf{\psi}_i^{\mathcal{K}} & l_i^{\mathcal{K}} \end{bmatrix}^{\top} \in \mathbb{R}^{d+1}$, which is then passed into the Cross-Correspondence Estimators. The input layer of these estimators are modified to account for the corresponding new input dimension.

The symmetry labels for each object are generated using an easily computed bisecting plane. 
% Now we will go into more details explaining how the symmetry labels are obtained.  
Given segmented point cloud demonstration of the gripper and the bottle/bowl in goal configuration, we apply Principle Component Analysis (PCA) to the individual object point cloud of the gripper and bottle/bowl. 

The symmetry labels for the gripper are defined using the PCA vector with largest principal component positioned at the gripper point cloud centroid, $\vec{\hat{s}}_{gripper}$. This vector defines the plane that bisects the gripper along its axis of actuation. 

For each point $\mathbf{p}_i^{gripper}$ in the gripper point cloud, compute the unit vector between it and gripper centroid $\mathbf{\mu}_{gripper}$, 

\begin{equation}
    \hat{\mathbf{v}}_i^{gripper} = \frac{\mathbf{p}_i^{gripper} - \mathbf{\mu}_{gripper}}{\|\mathbf{p}_i^{gripper} - \mathbf{\mu}_{gripper}\|} ,
\end{equation}

and retrieve the symmetry labels as the dot-product between the unit vectors,

\begin{equation}
l_i^{gripper} = \left\langle\vec{s}_{gripper}, \hat{\mathbf{v}}_{i}^{gripper}\right\rangle.
\end{equation}

To compute the symmetry labels for the bottle/bowl point clouds at training time, we retrieve the rotational symmetry axis, $\vec{\hat{v}}_{rot}$ of the bottle/bowl (pointing upward towards the bottle/bowl opening) using PCA. This vector is the largest principal component for the bottles and the smallest component for the bowls. We define a bisecting plane using both this symmetry axis, as well as the normalized vector pointing from the centroid of bottle/bowl to the centroid of the gripper, $\vec{\hat{v}}_{gripper}$. For the bottle points, the normal of the bisecting plane is found using the normalized cross-product of these two vectors, 
\begin{equation}
    \vec{s}_{bottle} = \frac{\vec{\hat{v}}_{rot} \times \vec{\hat{v}}_{gripper}}{\|\vec{\hat{v}}_{rot} \times \vec{\hat{v}}_{gripper}\|}, 
\end{equation}
which separates the bottle into left and right sides with respect to the gripper. For the bowl points, we orthoganalize the gripper vector, $\vec{\hat{v}}_{gripper}$ to symmetry vector, $\vec{\hat{v}}_{rot}$, 
\begin{equation}
    \vec{s}_{bowl} = \frac{\vec{\hat{v}}_{gripper} - \langle \vec{\hat{v}}_{gripper}, \vec{\hat{v}}_{rot} \rangle \vec{\hat{v}}_{rot}}{\| \vec{\hat{v}}_{gripper} - \langle \vec{\hat{v}}_{gripper}, \vec{\hat{v}}_{rot} \rangle \vec{\hat{v}}_{rot}\|}.
\end{equation}
This results in a bisecting plane that separates the bowl into a near and far half with respect to the gripper.
Similar to the gripper symmetry labels, the symmetry labels for the bottle/bowl are computed using the normalized vector between each bottle/bowl point and the bottle/bowl centroid, $\hat{\mathbf{v}}_i^{\{bottle, bowl\}}$, 
\begin{equation}
    l_i^{\{bottle, bowl\}} = \left\langle\vec{s}_{\{bottle, bowl\}}, \hat{\mathbf{v}}_i^{\{bottle, bowl\}}\right\rangle.
\end{equation}
At inference, instead of using the gripper location to compute $\vec{v}_{gripper}$ for the bottle and bowl labels, we use a random vector perpendicular to $\vec{v}_{rot}$. This allows us to use semantically meaningful symmetry labels at training time, and then arbitrarily break the symmetry at test time.

\begin{figure}[H]
\centering
\captionsetup[subfigure]{width=0.7\textwidth,justification=raggedright}
\begin{subfigure}{.5\textwidth}
  \centering
  \includegraphics[height=.36\linewidth]{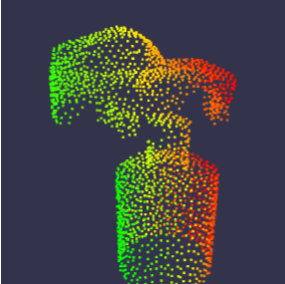}
  \caption{Symmetry labels for bottle and gripper}
  \label{fig:sym_bottle}
\end{subfigure}%
\begin{subfigure}{.5\textwidth}
  \centering
  \includegraphics[height=.36\linewidth]{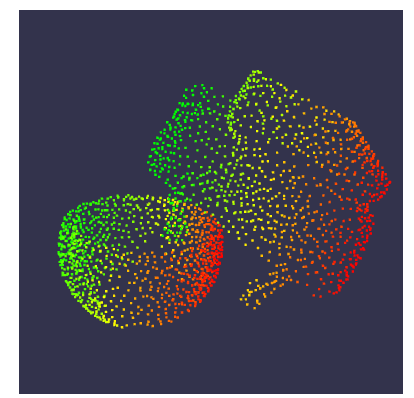}
  \caption{Symmetry labels for bowl and gripper}
  \label{fig:sym_bowl}
\end{subfigure}
\end{figure}

See Figure~\ref{fig:sym_bottle} and ~\ref{fig:sym_bowl} for visualization of symmetry labels obtained from aforementioned procedures, where the color spectrum of red represents symmetry labels of 1, and green represents -1. 

\subsubsection{Failure Cases}
\label{appendix:results-ndf-failure}

\begin{figure}[H]
\centering
\captionsetup[subfigure]{width=0.7\textwidth,justification=raggedright}
\begin{subfigure}{.5\textwidth}
  \centering
  \includegraphics[height=.36\linewidth]{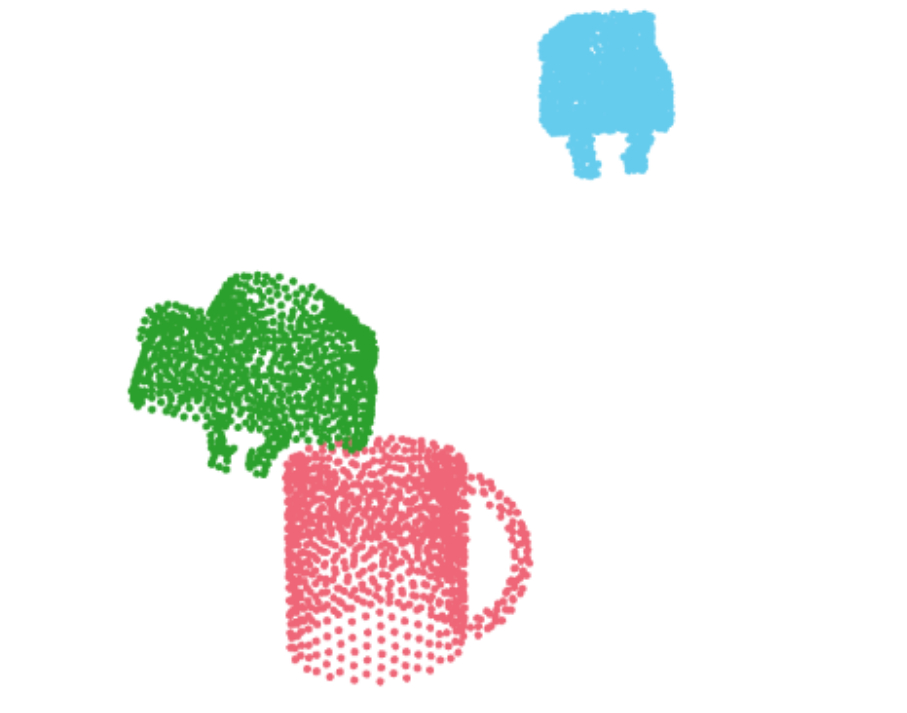}
  \caption{Failure of \textit{grasp} prediction. Predicted TAX-Pose for the gripper misses the rim of mug.}
  \label{fig:sub1_mug}
\end{subfigure}%
% \captionsetup[subfigure]{width=0.6\textwidth,justification=raggedright}
\begin{subfigure}{.5\textwidth}
  \centering
  \includegraphics[height=.36\linewidth]{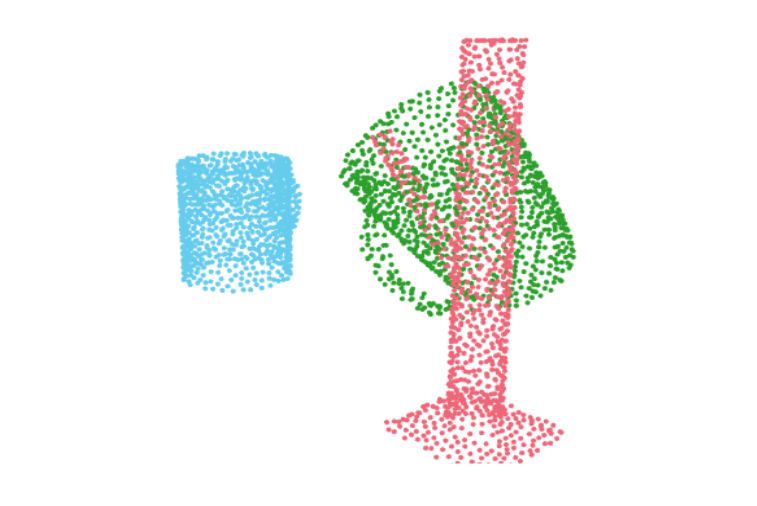}
  \caption{Failure of \textit{place} prediction. Predicted TAX-Pose for mug results in the mug handle misses the rack hanger completely.}
  \label{fig:sub2_mug}
\end{subfigure}
\caption{An illustration of unsuccessful TAX-Pose predictions for mug hanging. In both subfigures, red points represent the anchor object, blue points represent action object's starting pose, and green points represent action object's predicted pose.}
\label{fig:mug_fail}
\end{figure}

Some failure cases for TAX-Pose occur when the predicted gripper misses the rim of the mug by a xy-plane translation error, thus resulting in failure of grasp, as seen in Figure~\ref{fig:sub1_mug}. A common failure mode for the mug placement subtask is characterized by an erroneous transform prediction that results in the mug's handle completely missing the rack hanger, thus resulting in placement failure, as seen in Figure~\ref{fig:sub2_mug}. 

\subsection{PartNet-Mobility Tasks}
\label{appendix:results-pm}

\subsubsection{Expanded Results Tables}
\label{appendix:results-pm-full}

\begin{figure}[H]
    \centering
    \includegraphics[width=\textwidth]{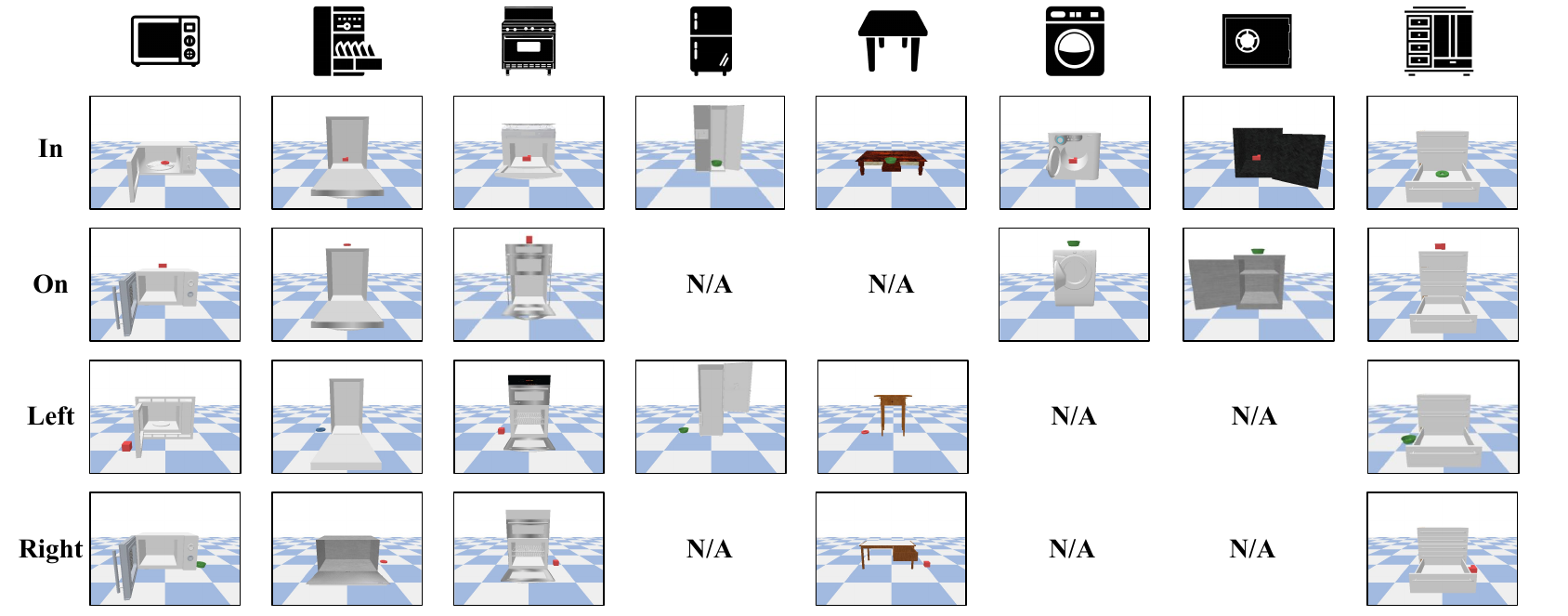}
    \caption{A visualization of all categories of anchor objects and associated semantic tasks, with action objects in ground-truth TAX-Poses used in simulation training.}
    \label{fig:allvis}
\end{figure}

In the main text, we presented aggregated results of the performance of each method by averaging the quantitative metrics for each sub-task for each object (``In'', ``On'', ``Left'', and ``Right'' in simulation and ``In'', ``On'' and ``Left'' in real-world), and then averaged across object classes to arrive at a single metric per method. Here, we present the per-class breakdown of performance. See Table \ref{tab:gi-Rt} for simulated results, and Table \ref{tab:rw_each} for real-world results.
\begin{table}[H]
\centering
\renewcommand{\arraystretch}{1.2}
\resizebox{\textwidth}{!}{
\setlength\tabcolsep{.2em}
\begin{tabular}{|lr|cc|cc|cc|cc|cc|cc|cc|cc|cc|}
\hline
 \rule{0pt}{2.5em}
 & \multicolumn{1}{l|}{} & \multicolumn{2}{c|}{{ \textbf{AVG.}}} & \multicolumn{2}{c|}{\clipart{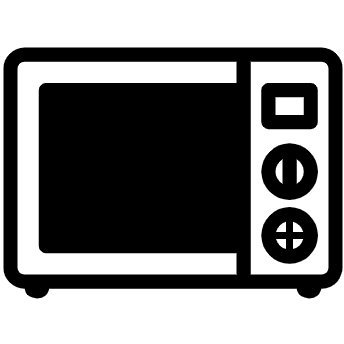}} &  \multicolumn{2}{c|}{\clipart{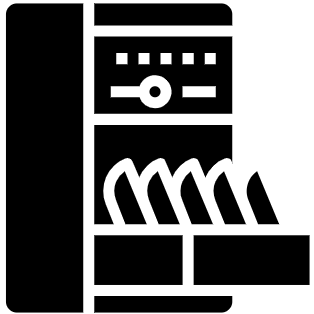}} & \multicolumn{2}{c|}{\clipart{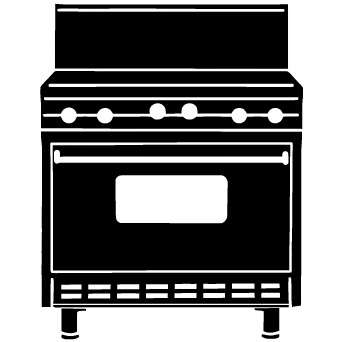}} & \multicolumn{2}{c|}{\clipart{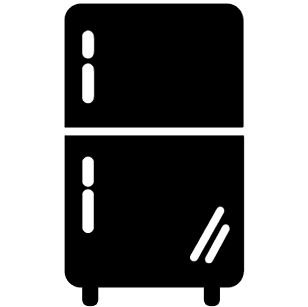}} & \multicolumn{2}{c|}{\clipart{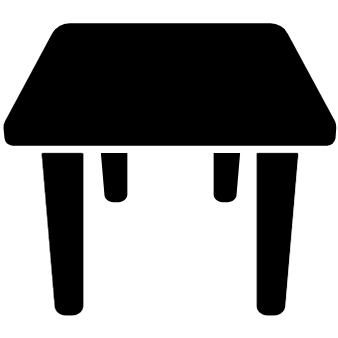}} & \multicolumn{2}{c|}{\clipart{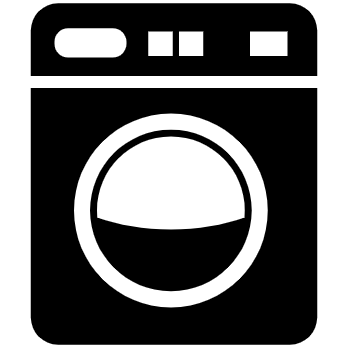}} & \multicolumn{2}{c|}{\clipart{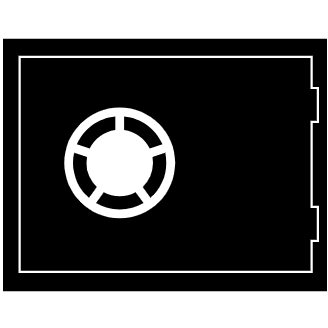}} & \multicolumn{2}{c|}{\clipart{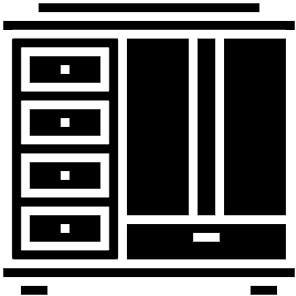}} \\ \cline{3-20} 
 & \multicolumn{1}{l|}{} & \textbf{$\mathcal{E}_\mathbf{R}$} & \textbf{$\mathcal{E}_\mathbf{t}$} & \textbf{$\mathcal{E}_\mathbf{R}$} & \textbf{$\mathcal{E}_\mathbf{t}$} & \textbf{$\mathcal{E}_\mathbf{R}$} & \textbf{$\mathcal{E}_\mathbf{t}$} & \textbf{$\mathcal{E}_\mathbf{R}$} & \textbf{$\mathcal{E}_\mathbf{t}$} & \textbf{$\mathcal{E}_\mathbf{R}$} & \textbf{$\mathcal{E}_\mathbf{t}$} & \textbf{$\mathcal{E}_\mathbf{R}$} & \textbf{$\mathcal{E}_\mathbf{t}$} & \textbf{$\mathcal{E}_\mathbf{R}$} & \textbf{$\mathcal{E}_\mathbf{t}$} & \textbf{$\mathcal{E}_\mathbf{R}$} & \textbf{$\mathcal{E}_\mathbf{t}$} & \textbf{$\mathcal{E}_\mathbf{R}$} & \textbf{$\mathcal{E}_\mathbf{t}$} \\ \hline
\multicolumn{1}{|l|}{\multirow{2}{*}{\textbf{Baselines}}} & E2E BC & 42.26 & 0.73 & 37.82 & 0.82 & 37.15 & 0.65 & 44.84 & 0.68 & 30.69 & 1.06 & 40.38 & 0.69 & 45.09 & 0.76 & 45.00 & 0.79 & 45.65 & 0.64 \\
\multicolumn{1}{|l|}{} & E2E DAgger~\cite{ross2011reduction} & 37.96 & 0.69 & 34.15 & 0.76 & 36.61 & 0.66 & 40.91 & 0.65 & 24.87 & 0.97 & 35.95 & 0.70 & 40.34 & 0.74 & 32.86 & 0.79 & 39.45 & 0.53 \\ \hline
\multicolumn{1}{|l|}{\multirow{2}{*}{\textbf{Ablations}}} & Traj. Flow~\cite{EisnerZhang2022FLOW} & 35.95 & 0.67 & 31.24 & 0.82 & 39.21 & 0.72 & 34.35 & 0.66 & 28.48 & 0.75 & 37.14 & 0.59 & 29.49 & 0.70 & 39.60 & 0.76 & 39.69 & 0.48 \\
\multicolumn{1}{|l|}{} & Goal Flow~\cite{EisnerZhang2022FLOW} & 26.64 & 0.17 & 25.88 & \textbf{0.15} & 25.05 & 0.15 & 30.62 & 0.15 & 27.61 & 0.10 & 28.01 & \textbf{0.18} & 20.96 & \textbf{0.24} & 29.02 & 0.23 & 22.13 & 0.20  \\
% \multicolumn{1}{|l|}{} & CPE Only & \multicolumn{1}{l}{} & \multicolumn{1}{l|}{} & \multicolumn{1}{l}{} & \multicolumn{1}{l|}{} & \multicolumn{1}{l}{} & \multicolumn{1}{l|}{} & \multicolumn{1}{l}{} & \multicolumn{1}{l|}{} & \multicolumn{1}{l}{} & \multicolumn{1}{l|}{} & \multicolumn{1}{l}{} & \multicolumn{1}{l|}{} & \multicolumn{1}{l}{} & \multicolumn{1}{l|}{} & \multicolumn{1}{l}{} & \multicolumn{1}{l|}{} & \multicolumn{1}{l}{} & \multicolumn{1}{l|}{} & \multicolumn{1}{l}{} & \multicolumn{1}{l|}{} \\ \hline
\hline
\multicolumn{1}{|l|}{\textbf{Ours}} & \textbf{TAX-Pose} & 6.64 & \textbf{0.16} & 6.85 & 0.16 & 2.05 & \textbf{0.10} & 3.87 & 0.12 & 4.04 & 0.08 & 12.71 & 0.31 & \textbf{6.87} & 0.37 & 5.89 & 0.13 & 14.93 & 0.18 \\ 
% \multicolumn{1}{|l|}{} & \textbf{TAX-Pose GC} & 7.74 & 0.17 & \textbf{4.43} &\textbf{0.13} & 3.27 & 0.13 & 4.16 & \textbf{0.12} & \textbf{3.3} & \textbf{0.08} & \textbf{10.76} & 0.26 & 7.36 & 0.30 & 6.28 & 0.14 & 22.37 & 0.21 \\ \hline

\multicolumn{1}{|l|}{} & \textbf{TAX-Pose GC} & \textbf{4.94} & \textbf{0.16} & \textbf{6.18} &0.16 & \textbf{1.75} & \textbf{0.10} & \textbf{2.94 }& \textbf{0.10} & \textbf{3.02} & \textbf{0.06} & \textbf{10.15} & 0.27 & 6.93 & 0.35 & \textbf{3.76} & \textbf{0.11} & \textbf{4.76} & \textbf{0.11} \\ \hline

\end{tabular}
}
\caption{Goal Inference Rotational and Translational Error Results ($\downarrow$). Rotational errors ($\mathcal{E}_\mathbf{R}$) are in degrees ($^\circ$) and translational errors ($\mathcal{E}_\mathbf{t}$) are in meters ($\mathrm{m}$). The lower the better.}
\label{tab:gi-Rt}
\end{table}
% \vspace{-25pt}
\begin{table}[H]
\renewcommand{\arraystretch}{1.2}
\resizebox{0.6\textwidth}{!}{
\setlength\tabcolsep{.2em}
\begin{tabular}{|l|lll|lll|lll|}
\hline
\multirow{2}{*}{} & \multicolumn{3}{c|}{\textbf{In}} & \multicolumn{3}{c|}{\textbf{On}} & \multicolumn{3}{c|}{\textbf{Left}} \\ \cline{2-10} 
 & \multicolumn{1}{l|}{\clipart{drawer}} & \multicolumn{1}{l|}{\clipart{fridge}} & \clipart{oven} & \multicolumn{1}{l|}{\clipart{drawer}} & \multicolumn{1}{l|}{\clipart{fridge}} & \clipart{oven} & \multicolumn{1}{l|}{\clipart{drawer}} & \multicolumn{1}{l|}{\clipart{fridge}} & \clipart{oven} \\ \hline
Goal Flow & \multicolumn{1}{l|}{0.00} & \multicolumn{1}{l|}{0.10} & 0.30 & \multicolumn{1}{l|}{0.05} & \multicolumn{1}{l|}{N/A} & 0.20 & \multicolumn{1}{l|}{0.50} & \multicolumn{1}{l|}{0.65} & 0.60 \\
\textbf{TAX-Pose} & \multicolumn{1}{l|}{\textbf{1.00}} & \multicolumn{1}{l|}{\textbf{1.00}} & \textbf{0.85} & \multicolumn{1}{l|}{\textbf{1.00}} & \multicolumn{1}{l|}{N/A} & \textbf{1.00} & \multicolumn{1}{l|}{\textbf{0.85}} & \multicolumn{1}{l|}{\textbf{0.90}} & \textbf{0.70} \\ \hline
\end{tabular}
}
\caption{Combined per-task results for real-world goal placement success rate.}
\label{tab:rw_each}
\end{table}
We further provide results per-sub-task in simulation. For each category of anchor objects, sub-tasks may or may not all be well-defined. For example, the doors of safes might occlude the action object completely in a demonstration for ``Left'' and ``Right'' tasks due to the handedness of the door; and a table's height might be too tall for the camera to see the action object placed during the ``Top'' task. To avoid these ill-defined cases, we omit object-category / sub-task pairings which cannot be consistently defined from training and evaluation. We show visualizations of each defined task for each object category in Figure \ref{fig:allvis}. Results for each sub-task can be found in Tables \ref{tab:gi-Rt_goal0}\footnote{Categories from left to right: microwave, dishwasher, oven, fridge, table, washing machine, safe, drawer.}, \ref{tab:gi-Rt_goal1}, \ref{tab:gi-Rt_goal2}, and \ref{tab:gi-Rt_goal3} respectively.

\begin{table}[H]
\centering
\renewcommand{\arraystretch}{1.2}
\resizebox{\textwidth}{!}{
\setlength\tabcolsep{.2em}
\begin{tabular}{|lr|cc|cc|cc|cc|cc|cc|cc|cc|cc|}
\hline
 \rule{0pt}{2.5em}
 & \multicolumn{1}{l|}{} & \multicolumn{2}{c|}{{ \textbf{AVG.}}} & \multicolumn{2}{c|}{\clipart{microwave}} &  \multicolumn{2}{c|}{\clipart{dishwasher}} & \multicolumn{2}{c|}{\clipart{oven}} & \multicolumn{2}{c|}{\clipart{fridge}} & \multicolumn{2}{c|}{\clipart{table}} & \multicolumn{2}{c|}{\clipart{washingmachine}} & \multicolumn{2}{c|}{\clipart{safe}} & \multicolumn{2}{c|}{\clipart{drawer}} \\ \cline{3-20} 
 & \multicolumn{1}{l|}{} & \textbf{$\mathcal{E}_\mathbf{R}$} & \textbf{$\mathcal{E}_\mathbf{t}$} & \textbf{$\mathcal{E}_\mathbf{R}$} & \textbf{$\mathcal{E}_\mathbf{t}$} & \textbf{$\mathcal{E}_\mathbf{R}$} & \textbf{$\mathcal{E}_\mathbf{t}$} & \textbf{$\mathcal{E}_\mathbf{R}$} & \textbf{$\mathcal{E}_\mathbf{t}$} & \textbf{$\mathcal{E}_\mathbf{R}$} & \textbf{$\mathcal{E}_\mathbf{t}$} & \textbf{$\mathcal{E}_\mathbf{R}$} & \textbf{$\mathcal{E}_\mathbf{t}$} & \textbf{$\mathcal{E}_\mathbf{R}$} & \textbf{$\mathcal{E}_\mathbf{t}$} & \textbf{$\mathcal{E}_\mathbf{R}$} & \textbf{$\mathcal{E}_\mathbf{t}$} & \textbf{$\mathcal{E}_\mathbf{R}$} & \textbf{$\mathcal{E}_\mathbf{t}$} \\ \hline
\multicolumn{1}{|l|}{\multirow{2}{*}{\textbf{Baselines}}} & E2E BC & 42.37 & 0.69 & 40.49 & 0.80 & 50.79 & 0.59 & 48.02 & 0.61 & 30.69 & 1.09 & 36.59 & 0.81 & 48.48 & 0.42 & 41.42 & 0.84 & 42.49 & 0.37 \\
\multicolumn{1}{|l|}{} & E2E DAgger~\cite{ross2011reduction} & 36.06 & 0.67 & 38.57 & 0.68 & 43.99 & 0.63 & 42.34 & 0.57 & 24.87 & 0.96 & 30.87 & 0.90 & 42.96 & 0.46 & 29.79 & 0.83 & 35.08 & 0.33 \\ \hline
\multicolumn{1}{|l|}{\multirow{2}{*}{\textbf{Ablations}}} & Traj. Flow~\cite{EisnerZhang2022FLOW} & 34.48 & 0.65 & 35.39 & 0.85 & 43.42 & 0.63 & 35.51 & 0.60 & 28.26 & 0.80 & 27.67 & 0.68 & 25.91 & 0.44 & 43.59 & 0.82 & 36.05 & 0.36 \\
\multicolumn{1}{|l|}{} & Goal Flow~\cite{EisnerZhang2022FLOW} & 27.49 & \textbf{0.21} & 25.41 & \textbf{0.08} & 31.07 & 0.13 & 27.05 & 0.27 & 27.80 & 0.11 & 29.02 & \textbf{0.38} & 19.22 & \textbf{0.36} & 31.56 & 0.18 & \textbf{28.81} & \textbf{0.19} \\
% \multicolumn{1}{|l|}{} & CPE Only & \multicolumn{1}{l}{} & \multicolumn{1}{l|}{} & \multicolumn{1}{l}{} & \multicolumn{1}{l|}{} & \multicolumn{1}{l}{} & \multicolumn{1}{l|}{} & \multicolumn{1}{l}{} & \multicolumn{1}{l|}{} & \multicolumn{1}{l}{} & \multicolumn{1}{l|}{} & \multicolumn{1}{l}{} & \multicolumn{1}{l|}{} & \multicolumn{1}{l}{} & \multicolumn{1}{l|}{} & \multicolumn{1}{l}{} & \multicolumn{1}{l|}{} & \multicolumn{1}{l}{} & \multicolumn{1}{l|}{} & \multicolumn{1}{l}{} & \multicolumn{1}{l|}{} \\ \hline
\hline
\multicolumn{1}{|l|}{\textbf{Ours}} & \textbf{TAX-Pose} & \textbf{11.74} & 0.23 & {\textbf{5.81}} & {0.11} & {\textbf{1.82}} & {\textbf{0.08}} & {\textbf{5.92}} & {\textbf{0.11}} & {\textbf{3.67}} & {\textbf{0.07}} & {\textbf{19.54}} & {0.41} & {\textbf{7.96}} & {0.63} & {\textbf{5.96}} & {\textbf{0.12}} & {43.27} & {0.33} \\ \hline
\end{tabular}
}
\caption{Goal Inference Rotational and Translational Error Results ($\downarrow$) for the ``\textbf{In'}' Goal. Rotational errors ($\mathcal{E}_\mathbf{R}$) are in degrees ($^\circ$) and translational errors ($\mathcal{E}_\mathbf{t}$) are in meters ($\mathrm{m}$). The lower the better.}
\label{tab:gi-Rt_goal0}
\end{table}
\begin{table}[H]
\centering
\renewcommand{\arraystretch}{1.2}
\resizebox{\textwidth}{!}{
\setlength\tabcolsep{.2em}
\begin{tabular}{|lr|cc|cc|cc|cc|cc|cc|cc|}
\hline
 \rule{0pt}{2.5em}
 & \multicolumn{1}{l|}{} & \multicolumn{2}{c|}{{ \textbf{AVG.}}} & \multicolumn{2}{c|}{\clipart{microwave}} &  \multicolumn{2}{c|}{\clipart{dishwasher}} & \multicolumn{2}{c|}{\clipart{oven}} & \multicolumn{2}{c|}{\clipart{washingmachine}} & \multicolumn{2}{c|}{\clipart{safe}} & \multicolumn{2}{c|}{\clipart{drawer}} \\ \cline{3-16} 
 & \multicolumn{1}{l|}{} & \textbf{$\mathcal{E}_\mathbf{R}$} & \textbf{$\mathcal{E}_\mathbf{t}$} & \textbf{$\mathcal{E}_\mathbf{R}$} & \textbf{$\mathcal{E}_\mathbf{t}$} & \textbf{$\mathcal{E}_\mathbf{R}$} & \textbf{$\mathcal{E}_\mathbf{t}$} & \textbf{$\mathcal{E}_\mathbf{R}$} & \textbf{$\mathcal{E}_\mathbf{t}$} & \textbf{$\mathcal{E}_\mathbf{R}$} & \textbf{$\mathcal{E}_\mathbf{t}$} & \textbf{$\mathcal{E}_\mathbf{R}$} & \textbf{$\mathcal{E}_\mathbf{t}$} & \textbf{$\mathcal{E}_\mathbf{R}$} & \textbf{$\mathcal{E}_\mathbf{t}$} \\ \hline
\multicolumn{1}{|l|}{\multirow{2}{*}{\textbf{Baselines}}} & E2E BC & 42.69 & 0.74 & 41.94 & 0.74 & 36.70 & 0.52 & 38.23 & 0.73 & 41.69 & 1.10 & 48.57 & 0.75 & 48.98 & 0.63 \\
\multicolumn{1}{|l|}{} & E2E DAgger~\cite{ross2011reduction} & 37.68 & 0.70 & 39.24 & 0.69 & 31.63 & 0.54 & 41.06 & 0.68 & 37.72 & 1.03 & 35.94 & 0.75 & 40.47 & 0.51 \\ \hline
\multicolumn{1}{|l|}{\multirow{2}{*}{\textbf{Ablations}}} & Traj. Flow~\cite{EisnerZhang2022FLOW} & 35.13 & 0.76 & 34.78 & 0.70 & 39.14 & 0.59 & 31.10 & 0.69 & 33.07 & 0.97 & 35.61 & 0.71 & 37.09 & 0.87 \\
\multicolumn{1}{|l|}{} & Goal Flow~\cite{EisnerZhang2022FLOW} & 22.10 & 0.20 & 27.82 & 0.26 & 20.43 & \textbf{0.09} & 34.66 & 0.10 & 22.71 & 0.12 & 26.48 & 0.27 & 0.48 & 0.32 \\
% \multicolumn{1}{|l|}{} & CPE Only & \multicolumn{1}{l}{} & \multicolumn{1}{l|}{} & \multicolumn{1}{l}{} & \multicolumn{1}{l|}{} & \multicolumn{1}{l}{} & \multicolumn{1}{l|}{} & \multicolumn{1}{l}{} & \multicolumn{1}{l|}{} & \multicolumn{1}{l}{} & \multicolumn{1}{l|}{} & \multicolumn{1}{l}{} & \multicolumn{1}{l|}{} & \multicolumn{1}{l}{} & \multicolumn{1}{l|}{} & \multicolumn{1}{l}{} & \multicolumn{1}{l|}{} & \multicolumn{1}{l}{} & \multicolumn{1}{l|}{} & \multicolumn{1}{l}{} & \multicolumn{1}{l|}{} \\ \hline
\hline
\multicolumn{1}{|l|}{\textbf{Ours}} & \textbf{TAX-Pose} & \textbf{4.45} & \textbf{0.12} & \textbf{4.21} & \textbf{0.12} & \textbf{2.29} & 0.10 & \textbf{2.73} & \textbf{0.09} & \textbf{5.77} & \textbf{0.10} & \textbf{5.81} & \textbf{0.13} & \textbf{5.89} & \textbf{0.19} \\ \hline
\end{tabular}
}
\caption{Goal Inference Rotational and Translational Error Results ($\downarrow$) for the ``\textbf{On}'' Goal. Rotational errors ($\mathcal{E}_\mathbf{R}$) are in degrees ($^\circ$) and translational errors ($\mathcal{E}_\mathbf{t}$) are in meters ($\mathrm{m}$). The lower the better.}
\label{tab:gi-Rt_goal1}
\end{table}
\begin{table}[H]
\centering
\renewcommand{\arraystretch}{1.2}
\resizebox{\textwidth}{!}{
\setlength\tabcolsep{.2em}
\begin{tabular}{|lr|cc|cc|cc|cc|cc|cc|cc|}
\hline
 \rule{0pt}{2.5em}
 & \multicolumn{1}{l|}{} & \multicolumn{2}{c|}{{ \textbf{AVG.}}} & \multicolumn{2}{c|}{\clipart{microwave}} &  \multicolumn{2}{c|}{\clipart{dishwasher}} & \multicolumn{2}{c|}{\clipart{oven}} & \multicolumn{2}{c|}{\clipart{fridge}} & \multicolumn{2}{c|}{\clipart{table}} & \multicolumn{2}{c|}{\clipart{drawer}} \\ \cline{3-16} 
 & \multicolumn{1}{l|}{} & \textbf{$\mathcal{E}_\mathbf{R}$} & \textbf{$\mathcal{E}_\mathbf{t}$} & \textbf{$\mathcal{E}_\mathbf{R}$} & \textbf{$\mathcal{E}_\mathbf{t}$} & \textbf{$\mathcal{E}_\mathbf{R}$} & \textbf{$\mathcal{E}_\mathbf{t}$} & \textbf{$\mathcal{E}_\mathbf{R}$} & \textbf{$\mathcal{E}_\mathbf{t}$} & \textbf{$\mathcal{E}_\mathbf{R}$} & \textbf{$\mathcal{E}_\mathbf{t}$} & \textbf{$\mathcal{E}_\mathbf{R}$} & \textbf{$\mathcal{E}_\mathbf{t}$} & \textbf{$\mathcal{E}_\mathbf{R}$} & \textbf{$\mathcal{E}_\mathbf{t}$} \\ \hline
\multicolumn{1}{|l|}{\multirow{2}{*}{\textbf{Baselines}}} & E2E BC & 44.87 & 0.74 & 30.95 & 0.89 & 36.86 & 0.72 & 56.86 & 0.52 & 34.35 & 1.03 & 31.69 & 0.77 & 46.86 & 0.78 \\
\multicolumn{1}{|l|}{} & E2E DAgger~\cite{ross2011reduction} & 41.32 & 0.68 & 31.40 & 0.84 & 38.49 & 0.73 & 47.64 & 0.51 & 36.47 & 0.99 & 27.72 & 0.73 & 39.83 & 0.51 \\ \hline
\multicolumn{1}{|l|}{\multirow{2}{*}{\textbf{Ablations}}} & Traj. Flow~\cite{EisnerZhang2022FLOW} & 38.85 & 0.58 & 31.87 & 1.07 & 39.48 & 0.44 & 39.48 & 0.44 & 28.71 & 0.69 & 41.06 & 0.73 & 40.70 & 0.31 \\
\multicolumn{1}{|l|}{} & Goal Flow~\cite{EisnerZhang2022FLOW} & 29.64 & \textbf{0.10} & 28.51 & \textbf{0.10} & 26.33 & \textbf{0.08} & 32.96 & \textbf{0.07} & 27.42 & 0.10 & 22.04 & \textbf{0.09} & 27.42 & 0.15  \\
% \multicolumn{1}{|l|}{} & CPE Only & \multicolumn{1}{l}{} & \multicolumn{1}{l|}{} & \multicolumn{1}{l}{} & \multicolumn{1}{l|}{} & \multicolumn{1}{l}{} & \multicolumn{1}{l|}{} & \multicolumn{1}{l}{} & \multicolumn{1}{l|}{} & \multicolumn{1}{l}{} & \multicolumn{1}{l|}{} & \multicolumn{1}{l}{} & \multicolumn{1}{l|}{} & \multicolumn{1}{l}{} & \multicolumn{1}{l|}{} & \multicolumn{1}{l}{} & \multicolumn{1}{l|}{} & \multicolumn{1}{l}{} & \multicolumn{1}{l|}{} & \multicolumn{1}{l}{} & \multicolumn{1}{l|}{} \\ \hline
\hline
\multicolumn{1}{|l|}{\textbf{Ours}} & \textbf{TAX-Pose} & \textbf{6.02} & 0.17 & \textbf{12.73} & 0.28 & \textbf{1.59} & 0.11 & \textbf{2.91} & 0.12 & \textbf{4.41} & \textbf{0.08} & \textbf{12.12} & 0.34 & \textbf{6.38} & \textbf{0.12} \\ \hline
\end{tabular}
}
\caption{Goal Inference Rotational and Translational Error Results ($\downarrow$) for the ``\textbf{Left}'' Goal. Rotational errors ($\mathcal{E}_\mathbf{R}$) are in degrees ($^\circ$) and translational errors ($\mathcal{E}_\mathbf{t}$) are in meters ($\mathrm{m}$). The lower the better.}
\label{tab:gi-Rt_goal2}
\end{table}
\begin{table}[H]
\centering
\renewcommand{\arraystretch}{1.2}
\resizebox{\textwidth}{!}{
\setlength\tabcolsep{.2em}
\begin{tabular}{|lr|cc|cc|cc|cc|cc|cc|}
\hline
 \rule{0pt}{2.5em}
 & \multicolumn{1}{l|}{} & \multicolumn{2}{c|}{{ \textbf{AVG.}}} & \multicolumn{2}{c|}{\clipart{microwave}} &  \multicolumn{2}{c|}{\clipart{dishwasher}} & \multicolumn{2}{c|}{\clipart{oven}} & \multicolumn{2}{c|}{\clipart{table}} & \multicolumn{2}{c|}{\clipart{drawer}} \\ \cline{3-14} 
 & \multicolumn{1}{l|}{} & \textbf{$\mathcal{E}_\mathbf{R}$} & \textbf{$\mathcal{E}_\mathbf{t}$} & \textbf{$\mathcal{E}_\mathbf{R}$} & \textbf{$\mathcal{E}_\mathbf{t}$} & \textbf{$\mathcal{E}_\mathbf{R}$} & \textbf{$\mathcal{E}_\mathbf{t}$} & \textbf{$\mathcal{E}_\mathbf{R}$} & \textbf{$\mathcal{E}_\mathbf{t}$} & \textbf{$\mathcal{E}_\mathbf{R}$} & \textbf{$\mathcal{E}_\mathbf{t}$} & \textbf{$\mathcal{E}_\mathbf{R}$} & \textbf{$\mathcal{E}_\mathbf{t}$} \\ \hline
\multicolumn{1}{|l|}{\multirow{2}{*}{\textbf{Baselines}}} & E2E BC & 39.11 & 0.76 & 37.89 & 0.86 & 24.26 & 0.77 & 36.27 & 0.88 & 52.86 & 0.48 & 44.26 & 0.78 \\
\multicolumn{1}{|l|}{} & E2E DAgger~\cite{ross2011reduction} & 36.80 & 0.73 & 27.40 & 0.84 & 32.31 & 0.74 & 32.61 & 0.82 & 49.27 & 0.46 & 42.40 & 0.78 \\ \hline
\multicolumn{1}{|l|}{\multirow{2}{*}{\textbf{Ablations}}} & Traj. Flow~\cite{EisnerZhang2022FLOW} & 35.33 & 0.71 & 22.93 & 0.66 & 34.78 & 1.22 & 31.29 & 0.92 & 42.71 & 0.37 & 44.93 & 0.36 \\
\multicolumn{1}{|l|}{} & Goal Flow~\cite{EisnerZhang2022FLOW} & 27.34 & 0.16 & 21.79 & 0.15 & 22.37 & 0.28 & 27.79 & 0.15 & 32.96 & \textbf{0.07} & 31.79 & 0.15  \\
% \multicolumn{1}{|l|}{} & CPE Only & \multicolumn{1}{l}{} & \multicolumn{1}{l|}{} & \multicolumn{1}{l}{} & \multicolumn{1}{l|}{} & \multicolumn{1}{l}{} & \multicolumn{1}{l|}{} & \multicolumn{1}{l}{} & \multicolumn{1}{l|}{} & \multicolumn{1}{l}{} & \multicolumn{1}{l|}{} & \multicolumn{1}{l}{} & \multicolumn{1}{l|}{} & \multicolumn{1}{l}{} & \multicolumn{1}{l|}{} & \multicolumn{1}{l}{} & \multicolumn{1}{l|}{} & \multicolumn{1}{l}{} & \multicolumn{1}{l|}{} & \multicolumn{1}{l}{} & \multicolumn{1}{l|}{} \\ \hline
\hline
\multicolumn{1}{|l|}{\textbf{Ours}} & \textbf{TAX-Pose} & \textbf{4.33} & \textbf{0.13} & \textbf{4.64} & \textbf{0.14} & \textbf{2.48} & \textbf{0.11} & \textbf{3.91} & \textbf{0.15} & \textbf{6.47} & 0.17 & \textbf{4.17} & \textbf{0.08} \\ \hline
\end{tabular}
}
\caption{Goal Inference Rotational and Translational Error Results ($\downarrow$) for the ``\textbf{Right}'' Goal. Rotational errors ($\mathcal{E}_\mathbf{R}$) are in degrees ($^\circ$) and translational errors ($\mathcal{E}_\mathbf{t}$) are in meters ($\mathrm{m}$). The lower the better.}
\label{tab:gi-Rt_goal3}
\end{table}

\subsubsection{Goal-Conditioned Variant}
\label{appendix:results-pm-gc}

We train our goal-conditioned variant \textbf{TAX-Pose GC} (Appendix \ref{appendix:training-architectures}) to predict the correct cross-pose across sub-tasks, incorporating a one-hot encoding of each sub-task (i.e. `top', `in', `left', 'right`) so the model can infer the desired semantic goal location. Importantly, as with the task-specific model (\textbf{TAX-Pose}), the \textbf{TAX-Pose GC} model is trained across \textbf{all PartNet-Mobility object categories}. We report the performance of the variants in Table \ref{tab:gi-Rt}.

\subsubsection{Failure Cases}
\label{appendix:results-pm-failure}
\begin{figure}[h]
\centering
\captionsetup[subfigure]{width=0.7\textwidth,justification=raggedright}
\begin{subfigure}{.5\textwidth}
  \centering
  \includegraphics[height=.36\linewidth]{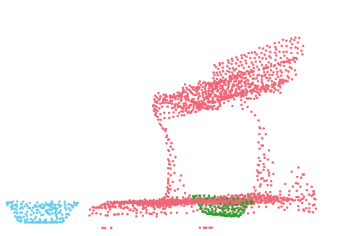}
  \caption{Failure of "In" prediction. Predicted TAX-Pose violates the physical constraints by penetrating the oven base too much.}
  \label{fig:sub1}
\end{subfigure}%
% \captionsetup[subfigure]{width=0.6\textwidth,justification=raggedright}
\begin{subfigure}{.5\textwidth}
  \centering
  \includegraphics[height=.36\linewidth]{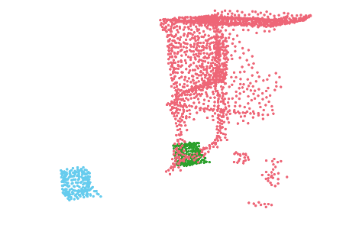}
  \caption{Failure of "Left" prediction. Predicted TAX-Pose violates the physical constraints by being in collision with the leg of the drawer.}
  \label{fig:sub2}
\end{subfigure}
\caption{An illustration of unsuccessful real-world TAX-Pose predictions. In both subfigures, red points represent the anchor object, blue points represent action object's starting pose, and green points represent action object's predicted pose.}
\label{fig:failure}
\end{figure}

Some failure cases for TAX-Pose occur when the predicted cross-pose does not respect the physical constraints in the scene. For example, as seen in Fig. \ref{fig:failure}. TAX-Pose would fail when the prediction violates the physical constraints of the objects, which in this case the action object collides with the anchor object. In the real world, this would yield the robot unable to plan a correct path.

% Failure illustration

\section{Task Details}
\label{appendix:taskdetails}

\subsection{NDF Task Details}
\label{appendix:taskdetails-ndf}

\begin{figure}[H]
    \centering
    \includegraphics[width=\textwidth]{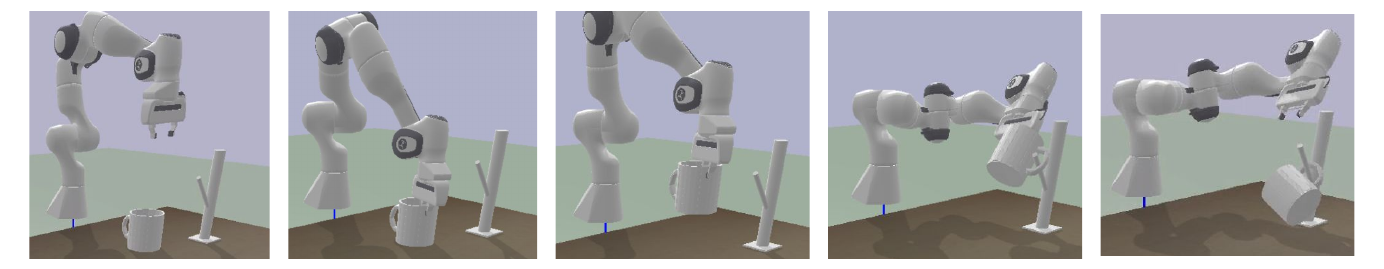}
    \caption{Visualization of Mug Hanging Task (Upright Pose). Mug hanging task is consisted of two stages, given a mug that is randomly initialized on the table, the model first predicts a $SE(3)$ transform from gripper end effector to the mug rim $\mathbf{T}_{g\rightarrow m}$, then grasp it by the rim. Next, the model predicts another $SE(3)$ transform from the mug to the rack $\mathbf{T}_{m\rightarrow r}$ such that the mug handle gets hanged on the the mug rack. }
    \label{fig:mug_task}
\end{figure}

In this section, we describe the Mug Hanging task of the NDF Tasks and experiments in detail. The Mug Hanging task is consisted of two sub tasks: \textit{grasp} and \textit{place}. A \textit{grasp} success is achieved when the mug is grasped stably by the gripper, while a \textit{place} success is achieved when the mug is hung stably on the hanger of the rack. Overall mug hanging success is achieved when the predicted transforms enable both grasp and place success for the same trial. See Figure~\ref{fig:mug_task} for a detailed breakdown of the mug hanging task in stages.
\subsubsection{Baseline Description}
\label{appendix:taskdetails-ndf-baselines}

In simulation, we compare our method to the results described in ~\cite{simeonov2021neural}. 
\begin{itemize}[nosep]
    \item \textbf{Dense Object Nets (DON)}~\cite{florence2018dense}: Using manually labeled semantic keypoints on the demonstration clouds, DON is used to compute sparse correspondences with the test objects. These correspondences are converted to a pose using SVD. A full description of usage of DON for the mug hanging task can be found in~\cite{simeonov2021neural}.
    \item \textbf{Neural Descriptor Field (NDF)}~\cite{simeonov2021neural}: Using the learned descriptor field for the mug, the positions of a constellation of task specific query points are optimized to best match the demonstration using gradient descent.
\end{itemize}
\subsubsection{Training Data}
\label{appendix:taskdetails-ndf-dataset}

To be directly comparable with the baselines we compared to, we use the exact same sets of demonstration data used to train the network in NDF~\cite{simeonov2021neural}, where the data are generated via teleportation in PyBullet, collected on 10 mug instances with random pose initialization. 

\subsubsection{Training and Inference}
\label{appendix:taskdetails-ndf-training}

Using the pretrained embedding network for mug and gripper, we train a grasping model for the grasping task to predict a transformation $\mathbf{T}_{g\rightarrow m}$ in gripper's frame from gripper to mug to complete the \textit{grasp} stage of the task. Similarly, using the pretrained embedding network for rack and mug, we train a placement model for the placing task to predict a transformation $\mathbf{T}_{m\rightarrow r}$ in mug's frame from mug to rack to complete the \textit{place} stage of the task. Both models are trained with the same combined loss $\mathcal{L}_{net}$ as described in the main paper. During inference, we simply use grasping model to predict the $\mathbf{T}_{g\rightarrow m}$ at test time, and placement model to predict $\mathbf{T}_{m\rightarrow r}$ at test time. 

\subsubsection{Motion Planning}
\label{appendix:taskdetails-ndf-motion}

After the model predicts a transformation $\mathbf{T}_{g\rightarrow m}$ and $\mathbf{T}_{m\rightarrow r}$, using the known gripper's world frame pose, we calculate the desired gripper end effector pose at grasping and placement, and pass the end effector to IKFast to get the desired joint positions of Franka at grasping and placement. Next we pass the desired joint positions at gripper's initial pose, and desired grasping joint positions to OpenRAVE motion planning library to solve for trajectory from gripper's initial pose to grasp pose, and then grasp pose to placement pose for the gripper's end effector. 
%~\ref{tab:mug_on_rack_few_shot}
%We find that our method is able to retain high success rates relatively better than the baselines with fewer demonstrations used.\newline
% \input{tables/mug_on_rack_few_shot}

\subsubsection{Real-World Experiments}
\label{appendix:taskdetails-ndf-real}

We pre-train the DGCNN embedding network with rotation-equivariant loss on ShapeNet mugs' simulated point clouds in simulation. Using the pre-trained embedding, we then train the full TAX-Pose model with the 10 collected real-world point clouds.
 
\subsection{PartNet-Mobility Object Placement Task Details}
\label{appendix:taskdetails-pm}

In this section, we describe the PartNet-Mobility Object Placement experiments in detail. We select a set of household furniture objects from the PartNet-Mobility dataset as the anchor objects, and a set of small rigid objects released with the Ravens simulation environment as the action objects. For each anchor object, we define a set of semantic goal positions (i.e. ‘top’, ‘left’, ‘right’, ‘in’), where action objects should be placed relative to each anchor. Each semantic goal position defines a unique task in our cross-pose prediction framework.
\subsubsection{Dataset Preparation}
\label{appendix:taskdetails-pm-dataset}

\textbf{Simulation Setup. } We leverage the PartNet-Mobility dataset \cite{Xiang2020-oz} to find common household objects as the anchor object for TAX-Pose prediction. The selected subset of the dataset contains 8 categories of objects. We split the objects into 54 seen and 14 unseen instances. During training, for a specific task of each of the seen objects, we generate an action-anchor objects pair by randomly sampling transformations from $SE(3)$ as cross-poses. The action object is chosen from the Ravens simulator's rigid body objects dataset \cite{Zeng2020-tk}. We define a subset of four tasks (``In'', ``On'', ``Left'' and ``Right'') for each selected anchor object.  
Thus, there exists a ground-truth cross-pose (defined by human manually) associated with each defined specific task. We then use the ground-truth TAX-Poses to supervise each task's TAX-Pose prediction model. For each observation action-anchor objects pair, we sample 100 times using the aforementioned procedure for the training and testing datasets. 

\textbf{Real-World Setup. }In real-world, we select a set of anchor objects: Drawer, Fridge, and Oven and a set of action objects: Block and Bowl. We test 3 (``In'', ``On'', and ``Left'') TAX-Pose models in real-world without retraining or finetuning. The point here is to show the method capability of generalizing to unseen real-world objects.

\subsubsection{Metrics}
\label{appendix:taskdetails-pm-metrics}

\textbf{Simulation Metrics. }In simulation, with access to the object's ground-truth pose, we are able to quantitatively calculate translational and rotation error of the TAX-Pose prediction models. Thus, we report the following metrics on a held-out set of anchor objects in simulation:

\begin{tabular}{@{}p{.48\textwidth}p{.48\textwidth}@{}} % Remove margins left and right.
     \textit{Translational Error}: The L2 distance between the inferred cross-pose translation ($\mathbf{t}_{\mathcal{A}\mathcal{B}}^\mathrm{pred}$) and the ground-truth pose translation ($\mathbf{t}_{\mathcal{A}\mathcal{B}}^\mathrm{GT}$). & 
     \textit{Rotational Error}: The geodesic $SO(3)$ distance \cite{huynh2009metrics, hartley2013rotation} between the predicted cross-pose rotation ($\mathbf{R}_{\mathcal{A}\mathcal{B}}^\mathrm{pred}$) and the ground-truth rotation ($\mathbf{R}_{\mathcal{A}\mathcal{B}}^\mathrm{GT}$).
     \\
     \parbox{0.5\textwidth}{\begin{equation*} 
        \mathcal{E}_\mathbf{t} = ||\mathbf{t}_{\mathcal{A}\mathcal{B}}^\mathrm{pred} - \mathbf{t}_{\mathcal{A}\mathcal{B}}^\mathrm{GT}||_2
    \end{equation*}} 
     & 
     \parbox{0.5\textwidth}{\begin{equation*} 
        \mathcal{E}_\mathbf{R} = \frac{1}{2} \arccos{\left(\frac{\mathrm{tr}(\mathbf{R}_{\mathcal{A}\mathcal{B}}^{\mathrm{pred}\top} \mathbf{R}_{\mathcal{A}\mathcal{B}}^\mathrm{GT})-1}{2}\right)}
    \end{equation*}}
\end{tabular}

\textbf{Real-World Metrics. }In real-world, due to the difficulty of defining ground-truth TAX-Pose, we instead manually, qualitatively define goal ``regions'' for each of the anchor-action pairs. The goal-region should have the following properties:
\begin{itemize}[noitemsep]
    \item The predicted TAX-Pose of the action object should appear visually correct. For example, if the specified task is ``In'', then the action object should be indeed contained within the anchor object after being transformed by predicted TAX-Pose.
    \item The predicted TAX-Pose of the action object should not violate physical constraints of the workspace and of the relation between the action and anchor objects. Specifically, the action object should not interfere with/collide with the anchor object after being transformed by the predicted TAX-Pose. See Figure \ref{fig:failure} for an illustration of TAX-Pose predictions that fail to meet this criterion. 
\end{itemize}

\subsubsection{Motion Planning}
\label{appendix:taskdetails-pm-motion}

In both simulated and real-world experiments, we use off-the-shelf motion-planning tools to find a path between the starting pose and goal pose of the action object.

\textbf{Simulation. }To actuate the action object from its starting pose $\mathbf{T}_0$ to its goal pose transformed by the predicted TAX-Pose $\hat{\mathbf{T}}_{\mathcal{A}\mathcal{B}}\mathbf{T}_0$, we plan a path free of collision. Learning-based methods such as \cite{danielczuk2021object} deal with collision checking with point clouds by training a collision classifier. A more data-efficient method is by leveraging computer graphics techniques, transforming the point clouds into marching cubes \cite{lorensen1987marching}, which can then be used to efficiently reconstruct meshes. Once the triangular meshes are reconstructed, we can deploy off-the-shelf collision checking methods such as FCL \cite{pan2012fcl} to detect collisions in the planned path. Thus, in our case, we use position control to plan a trajectory of the action object $\mathcal{A}$ to move it from its starting pose to the predicted goal pose. We use OMPL \cite{sucan2012open} as the motion planning tool and the constraint function passed into the motion planner is from the output of FCL after converting the point clouds to meshes via marching cubes. 

\begin{figure}
    \centering
    \includegraphics[width=0.6\textwidth]{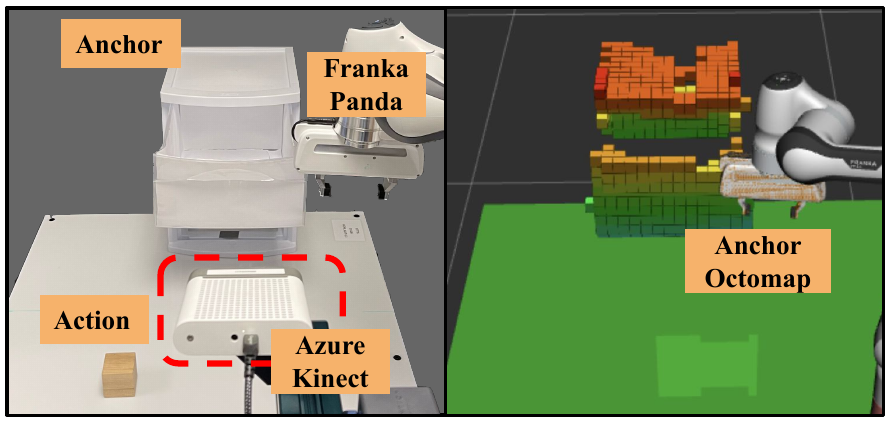}
    \captionof{figure}{Real-world experiments illustration. \textbf{Left: }work-space setup for physical experiments. \textbf{Center: }Octomap visualization of the perceived anchor object.} 
    \label{fig:image}
\end{figure}%

\textbf{Real World. }In real-world experiments, we need to resolve several practical issues to make TAX-Pose prediction model viable. First, we do not have access to a mask that labels action and anchor objects. Thus, we manually define a mask by using a threshold value of $y$-coordinate to automatically detect discontinuity in $y$-coordinates, representing the gap of spacing between action and anchor objects upon placement. Next, grasping action objects is a non-trivial task. Since, we are only using 2 action objects (a cube and a bowl), we manually define a grasping primitive for each action object. This is done by hand-picking an offset from the centroid of the action object before grasping, and an approach direction after the robot reaches the pre-grasp pose to make contacts with the object of interest. The offsets are chosen via kinesthetic teaching on the robot when the action object is under identity rotation (canonical pose). Finally, we need to make an estimation of the action's starting pose for motion planning. This is done by first statistically cleaning the point cloud~\cite{EisnerZhang2022FLOW} of the action object, and then calculating the centroid of the action object point cloud as the starting position. For starting rotation, we make sure the range of the rotation is not too large for the pre-defined grasping primitive to handle. Another implementation choice here is to use ICP~\cite{besl1992method} calculate a transformation between the current point cloud to a pre-scanned point cloud in canonical (identity) pose. We use the estimated starting pose to guide the pre-defined grasp primitive. Once a successful grasp is made, the robot end-effector is rigidly attached to the action object, and we can then use the same predicted TAX-Pose to calculate the end pose of the robot end effector, and thus feed the two poses into MoveIt! to get a full trajectory in joint space. Note here that the collision function in motion planning is comprised of two parts: workspace and anchor object. That is, we first reconstruct the workspace using boxes to avoid collision with the table top and camera mount, and we then reconstruct the anchor object in RViz using Octomap~\cite{hornung2013octomap} using the cleaned anchor object point cloud. In this way, the robot is able to avoid collision with the anchor object as well. See Figure \ref{fig:image} for the workspace.

\subsubsection{Baselines Description}
\label{appendix:taskdetails-pm-baselines}

In simulation, we compare our method to a variety of baseline methods.
\begin{itemize}
    \item \textbf{E2E Behavioral Cloning}: Generate motion-planned trajectories using OMPL that take the action object from start to goal. These serve as ``expert'' trajectories for Behavioral Cloning (BC). Our policy is represented as a PointNet++~\cite{qi2017pointnet++} network that, at each time step, takes as input the point cloud observation of the action and anchor objects and outputs an incremental 6-DOF transformation that imitates the expert trajectory. The 6-DoF transformation is expressed using Euclidean $xyz$ translation and rotation quaternion. The final achieved pose of the action object at the terminal state is used for computing the evaluation metrics.

    \item \textbf{E2E DAgger}: Using the same BC dataset and the same PointNet++~\cite{qi2017pointnet++} architecture as above, we train a policy that outputs the same transformation representation as in BC using DAgger~\cite{ross2011reduction}. The final achieved pose of the action object at the terminal state is used for computing the evaluation metrics. 
    
    \item \textbf{Trajectory Flow}: Using the same BC dataset with DAgger, we train a dense policy using PointNet++~\cite{qi2017pointnet++} to predict a dense per-point 3D flow vector at each time step instead of a single  6-DOF transformation. Given this dense per-point flow, we add the per-point flow to each point of the current time-step's point cloud, and we are able to extract a rigid transformation between the current point cloud and the point cloud transformed by adding per-point flow vectors using SVD, yielding the next pose. The  final achieved pose of the action object at the terminal state is used for computing the evaluation metrics.
    
    \item \textbf{Goal Flow}: Instead of training a multi-step policy to reach the goal, we train a PointNet++~\cite{qi2017pointnet++} network to output a single dense flow prediction which assigns a per-point 3D flow vector that points from each action object point from its starting pose directly to its corresponding goal location. Given this dense per-point flow, we add the per-point flow to each point of the start point cloud, and we are able to extract a rigid transformation between the start point cloud and the point cloud transformed by adding per-point flow vectors using SVD, yielding goal pose. We pass the start and goal pose into a motion planner (OMPL) and execute the planned trajectory. The final achieved pose of the action object at the terminal state is used for computing the evaluation metrics.
\end{itemize}

\section{Author Contributions}
\label{appendix:contributions}

\underline{Chuer Pan} designed and implemented the cross-correspondence TAX-Pose model, correspondence residuals, and the bi-directional weighted SVD method. She also designed, implemented, and conducted the simulation experiments of the NDF tasks as well as the study on varying the number of demos and the various ablation experiments. She also designed, implemented, and trained the models for mug hanging task in the real-world.

\underline{Brian Okorn} proposed, designed, and implemented the cross-correspondence model, correspondence residuals, and the bi-directional weighted SVD method, as well as analyzed the invariances of the current framework.

\underline{Harry Zhang} wrote early prototypes of TAX-Pose using residual flows without cross-attention, which later became a baseline of TAX-Pose. He also wrote the infrastructure of the PartNet-Mobility object placement task's simulated data collection and training. He also wrote all the baselines in the PartNet-Mobility placement task. He designed and conducted all real-world robot experiments for both tasks in the real-world.

\underline{Ben Eisner} designed the PartNet-Mobility tasks, implemented the Pybullet physics environment required for their rendering and simulation, and conducted training and evaluation of TAX-Pose on the PartNet-Mobility dataset. He also designed and evaluated the goal-conditioned variant of TAX-Pose.

% \bibliography{ref}  % .bib
% \putbib{ref}
% \end{document}
% \putbib[ref2]
% \end{bibunit}
\end{document}